%% file: paper_arxiv_version.tex
\documentclass[nonacm,acmtog]{acmart}
\AtBeginDocument{%
  }

\acmJournal{TOG}

\usepackage{multirow}
\usepackage{booktabs}
\begin{document}


\title{MegaParts: Scaling Part-Aware 3D Object Generation to 300 Parts via Token-Efficient Autoregressive Modeling}

\author{Manwen Liao}
\authornote{Equal contribution.}
\authornote{Work done during internship at Shanghai Artificial Intelligence Laboratory.}
\orcid{0009-0003-5944-1556}
\affiliation{%
  \institution{The University of Hong Kong},
  \institution{Shanghai Artificial Intelligence Laboratory}
  \country{China}
}
\email{manwen@connect.hku.hk}

\author{Xinyu Lian}
\orcid{0009-0002-0735-1308}
\authornotemark[1]
\authornotemark[2]
\affiliation{%
  \institution{Fudan University},
  \institution{Shanghai Artificial Intelligence Laboratory}
  \country{China}
}
\email{xylian25@m.fudan.edu.cn}

\author{Jian Mao}
\orcid{0009-0002-1979-5838}
\authornotemark[1]
\authornotemark[2]
\affiliation{%
  \institution{Tongji University},
  \institution{Shanghai Artificial Intelligence Laboratory}
  \country{China}
}
\email{2211108@tongji.edu.cn}

\author{Kaixu Chen}
\orcid{0009-0003-7700-5432}
\affiliation{%
  \institution{University of Science and Technology of China},
  \institution{Shanghai Artificial Intelligence Laboratory}
  \country{China}
}
\email{chenkaixu.me@outlook.com}

\author{Li Luo}
\orcid{0009-0007-7766-9166}
\affiliation{%
  \institution{The University of Hong Kong},
  \institution{Shanghai Artificial Intelligence Laboratory}
  \country{China}
}
\email{u3014135@connect.hku.hk}

\author{Jinghao Yan}
\orcid{0009-0003-9720-0681}
\affiliation{%
  \institution{Tongji University},
  \institution{Shanghai Artificial Intelligence Laboratory}
  \country{China}
}
\email{jinghaoyan943@gmail.com}

\author{Wanshui Gan}
\orcid{0000-0002-6720-6500}
\affiliation{%
  \institution{Shanghai Artificial Intelligence Laboratory}
  \country{China}
}
\email{wanshuigan@gmail.com}

\author{Qiao Yu}
\orcid{0000-0002-6392-9461}
\affiliation{%
  \institution{Shanghai Artificial Intelligence Laboratory}
  \country{China}
}
\email{yuqiao0303@gmail.com}

\author{Weitian Zhang}
\orcid{0000-0001-6760-1132}
\affiliation{%
  \institution{Shanghai Jiao Tong University}
  \country{China}
}
\email{weitianzhang@sjtu.edu.cn}

\author{Chunhua Shen}
\orcid{0000-0002-8648-8718}
\affiliation{%
  \institution{Shanghai Artificial Intelligence Laboratory},
  \institution{Zhejiang University}
  \country{China}
}
\email{chhshen@gmail.com}

\author{Guang Chen}
\orcid{0000-0002-7416-592X}
\authornote{Corresponding author.}
\affiliation{%
  \institution{Tongji University},
  \institution{Shanghai Innovation Institute}
  \country{China}
}
\email{guangchen@tongji.edu.cn}

\author{Bo Dai}
\orcid{0000-0003-0777-9232}
\authornotemark[3]
\affiliation{%
  \institution{The University of Hong Kong}
  \country{China}
}
\email{doubledaibo@gmail.com}

\author{Xudong Xu}
\orcid{0009-0003-8858-0918}
\affiliation{%
  \institution{Shanghai Artificial Intelligence Laboratory}
  \country{China}
}
\email{xudongxu9710@gmail.com}

\author{Zhaoyang Lyu}
\orcid{0009-0002-7657-0221}
\authornotemark[3]
\affiliation{%
  \institution{Shanghai Artificial Intelligence Laboratory}
  \country{China}
}
\email{lyuzhaoyang@link.cuhk.edu.hk}










\newcommand{\lyu}[1]{\textcolor{red}{zhaoyang: #1}}
\colorlet{red}{black}
\begin{abstract}
  Part-aware 3D object generation is essential for graphics applications such as controllable modeling, editing, and articulation, where objects are represented as coherent assemblies of semantic parts. 
  However, existing part-aware generation methods, do not scale well to highly complex objects. 
  As the number of parts increases, generating detailed geometry becomes prohibitively expensive in token length and memory. 
  We introduce MegaParts, a scalable autoregressive 3D generation framework to address this challenge by combining structured sequence modeling with a token-efficient vector-quantized shape tokenizer. 
  Our tokenizer learns discrete latent representations for part-level geometry by minimizing token usage subject to high-fidelity reconstruction, enabling adaptive-length tokenization based on geometric complexity. 
  On top of this compact representation, we train a large language model to generate object bounding boxes, part bounding boxes, and part shape tokens within a unified structured sequence. Combined with efficient long-context training strategy, our token-efficient formulation scales to objects with up to 300 parts and sequence lengths up to 256k tokens. This substantially extends the scale of part-aware 3D generation while preserving compositional structure and enabling fine-grained part-level control. Our method achieves higher mesh quality than baseline autoregressive and diffusion models, showing that compressed discrete part tokens improve not only scalability but also the achievable fidelity of generated geometry. These results suggest that LLM native token-efficient autoregressive modeling is a compelling alternative to diffusion for large-scale part-aware 3D generation. The project page is available at \href{https://expmaster.github.io/megaparts_webpage}{https://expmaster.github.io/megaparts\_webpage}. 
\end{abstract}

\begin{teaserfigure}
\includegraphics[width=\textwidth]{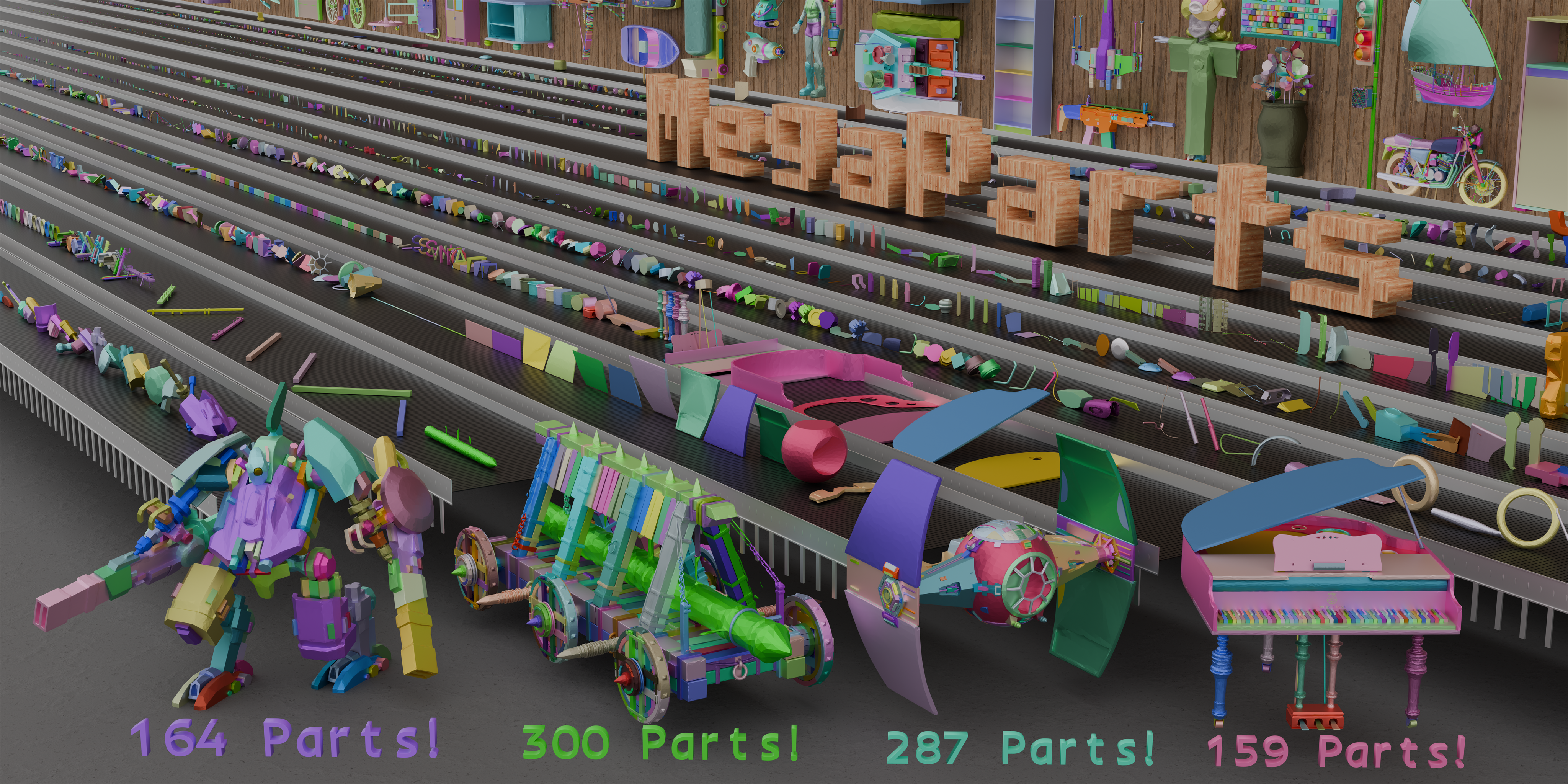}
\caption{MegaParts can generate complex objects with up to 300 parts. Like factory assembly lines, it autoregressively manufactures part meshes and then assembles them into complete 3D objects, substantially extending the scope and capability of part-aware 3D generation while maintaining geometry quality.}
\Description{figure description}
\end{teaserfigure}


\maketitle

\makeatletter
\fancypagestyle{standardpagestyle}{%
  \fancyhf{}%
  \fancyhead[LE,RO]{\@headfootfont\thepage}%
  \renewcommand{\headrulewidth}{0pt}%
  \renewcommand{\footrulewidth}{0pt}%
}
\pagestyle{standardpagestyle}
\makeatother

\section{Introduction}

\input{subsections/introduction}

\section{Related Work}
\input{subsections/related_work}

\section{Methodology}
\label{sec:vae}
\input{subsections/3_1_vqvae}
\input{subsections/3_2_representation_training}
\input{subsections/3_3_token_count}
\input{subsections/3_4_autoregressive}

\section{Experimental Results}
\input{subsections/6_1_experiment_intro}

\input{subsections/7_1_ablation_intro}

\section{Applications}

\input{subsections/8_1_detail_second_time_creation}
\input{subsections/8_2_application_articultation_creation}

\section{Conclusions and Discussions}

\input{subsections/9_1_c_d}

\bibliographystyle{ACM-Reference-Format}
\bibliography{sample-base}

\begin{figure*}[htp]
  \centering
  \includegraphics[width=0.95\linewidth]{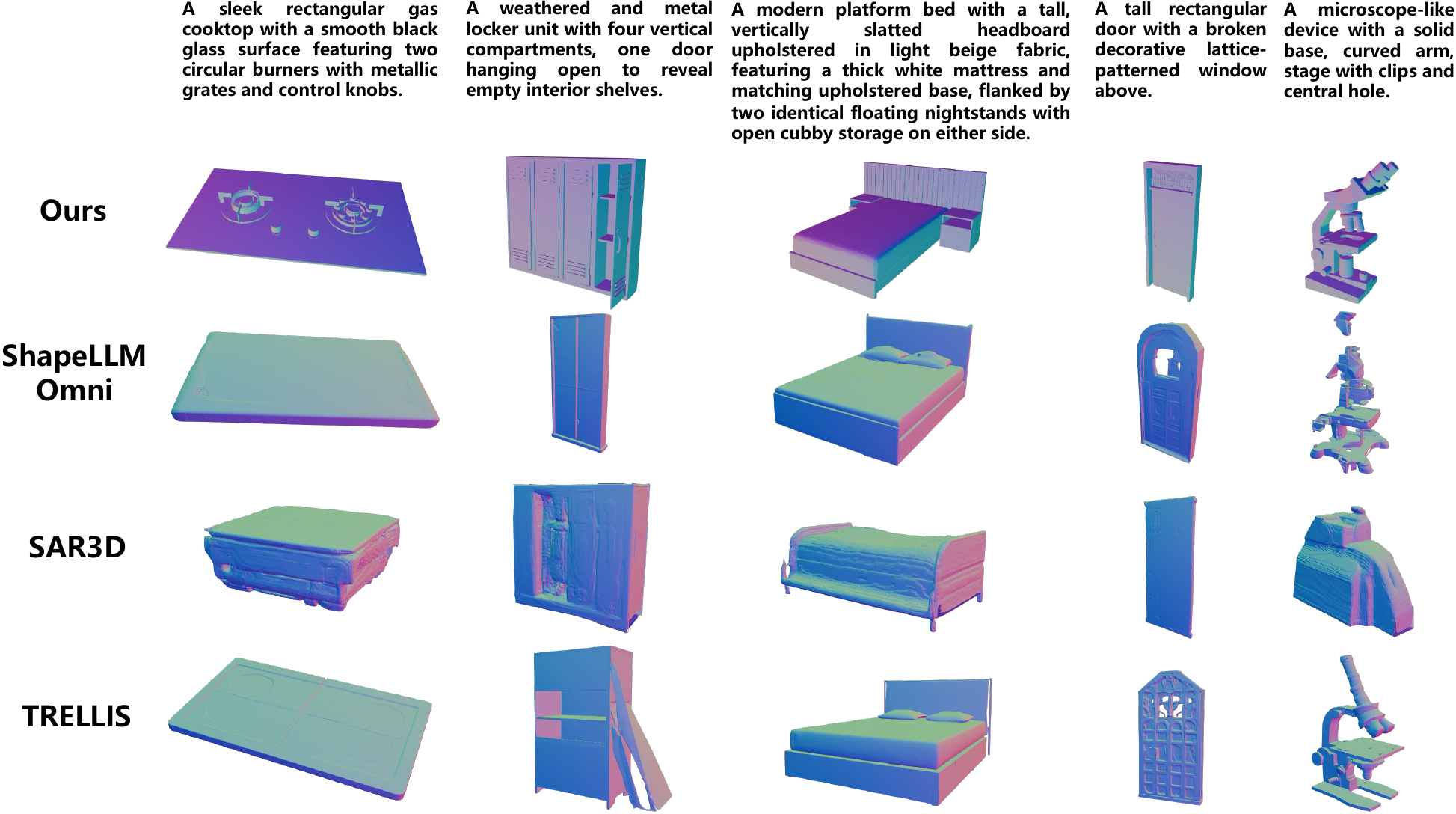}
  \caption{Qualitative comparison between our text conditioned generation results against baselines. Our results show superior alignment with given text prompt, demonstrating our model's strong prompt following ability.}
  \label{fig:text_cond}
\end{figure*}

\begin{figure*}[htp]
  \centering
  \includegraphics[width=0.95\linewidth]{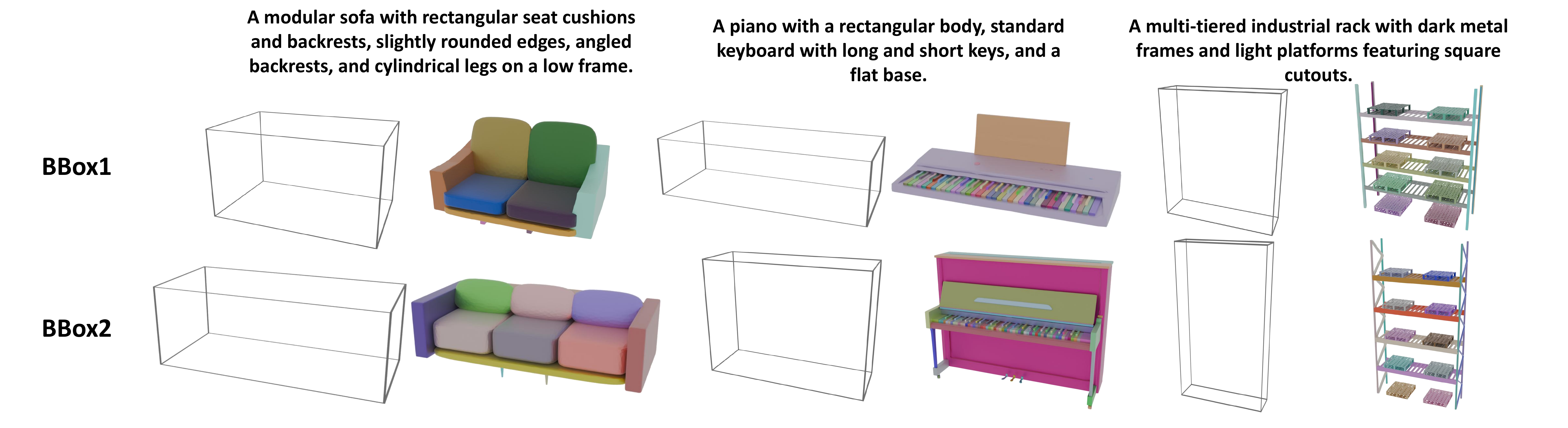}
  \caption{Results of global bounding-box-conditioned generation. As the input bounding-box configuration changes, our model automatically synthesizes corresponding part layouts and meshes. The generated results align closely with the prescribed bounding boxes, demonstrating effective structural control.
  }
  \label{fig:bbox}
\end{figure*}

\begin{figure*}[htp]
  \centering
  
  \includegraphics[width=0.85\linewidth]{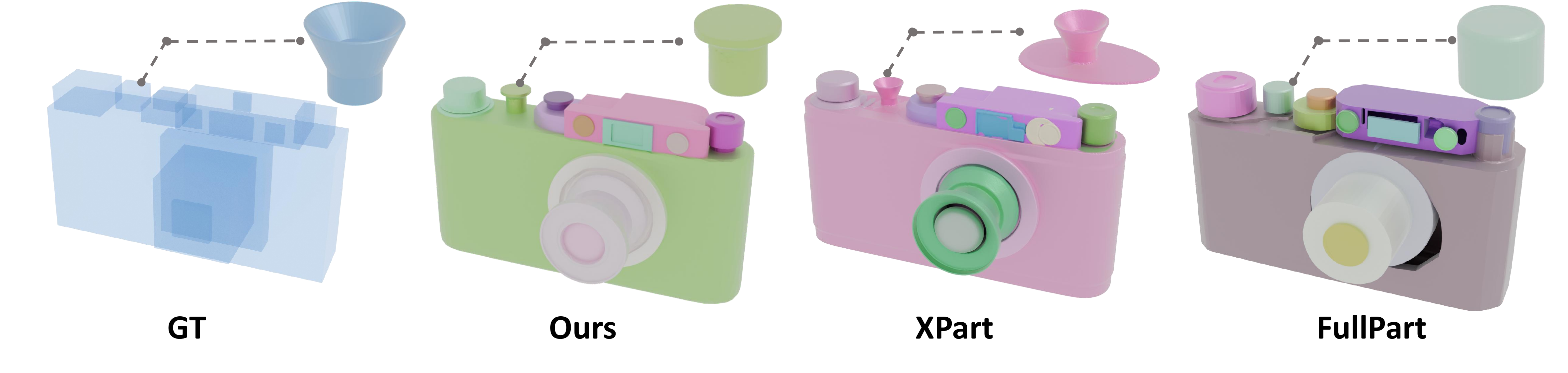}
  \caption{Results of part bboxes conditioned generation. Our model {\color{red} produces} semantically reasonable part meshes with high quality geometry. 
  }
  \label{fig:part_cond}
\end{figure*}

\begin{figure*}[htp]
  \centering
  \includegraphics[width=0.90\linewidth]{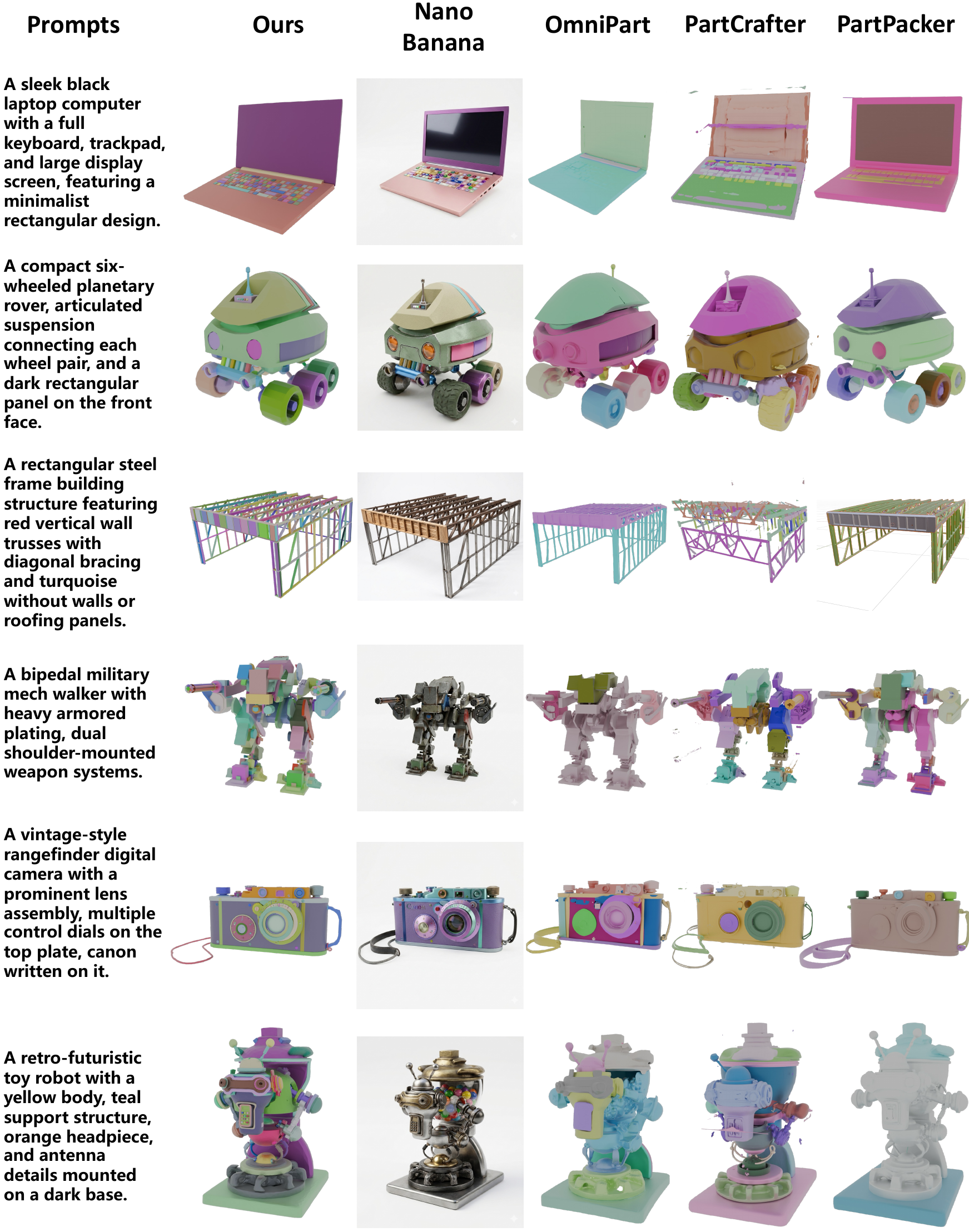}
  \caption{Comparison between our results with image conditioned part-aware generation methods. We use Nano Banana to refine rendered images of our text conditioned generation results and use them as input to baselines. Our resulted geometry surpass baselines and our part segmentations are much more structured and detailed than other methods.}
  \label{fig:part_aware}
\end{figure*}

\clearpage
\appendix



\section{Implementation Details}
\input{subsections/a_1_1_model_sepc}

\input{subsections/a_1_2_data_collection_preprocess}
\input{subsections/a_1_3_training_recipes}
\input{subsections/a_2_ablations}
\input{subsections/a_4_more_experiments}

\input{subsections/a_3_model_capability}

\end{document}

%% file: subsections/introduction.tex
3D generation has attracted increasing attention in recent years due to its broad applications in VR/AR, games and embodied simulation. Existing holistic 3D generation methods~\cite{10.1145/3592442,zhao2023michelangelo,zhao2025hunyuan3d20scalingdiffusion,10.1145/3658146,10.1145/3730841,li2025craftsman3dhighfidelitymeshgeneration,xiang2024structured,wu2025direct3ds2gigascale3dgeneration,xiang2025trellis2,chen2025ultra3defficienthighfidelity3d,lai2025latticedemocratizehighfidelity3d,sam3dteam2025sam3d3dfyimages,ye2025shapellm,10.1145/3757377.3763812} have made remarkable progress in producing high-quality complete shapes. However, they typically model an object as a single entity and ignore its compositional nature. This makes it difficult to control { \color{red}structural} components, perform localized editing, or support articulation. By explicitly representing objects as structured assemblies of {\color{red} structural} components, part-aware generation~\cite{lin2025partcrafter,tang2024partpacker,yang2025holopart,10.1145/3730840,10.1145/3757377.3763872,yan2025xparthighfidelitystructure,ding2025fullpart,wang2025partxmllmpartaware3dmultimodal,dai2025meshcoder,chen2025autopartgenautogressive3dgeneration} offers a more natural way for controllable 3D generation, editing and articulation. Moreover, recent studies~\cite{lai2025latticedemocratizehighfidelity3d,yang2025log3dultrahighresolution3dshape} suggest that structured and localizable representations are particularly beneficial for scalable 3D generation, which naturally aligns with part-aware formulation. As a result, part-aware generation provides a promising foundation for scalable and controllable 3D content creation.

Current part-aware generation pipelines~\cite{yang2025holopart,tang2024partpacker,yan2025xparthighfidelitystructure,lin2025partcrafter,10.1145/3730840,chen2025autopartgenautogressive3dgeneration, ding2025fullpart,wang2025partxmllmpartaware3dmultimodal} typically leverage diffusion-based architectures, relying on predefined segmentation results or denoised object layouts to guide local geometry generation. 
Despite their progress, these methods inherit scalability limitations from diffusion models. Experiments in image generation demonstrate that naive diffusion models often struggle as the token number increases~\cite{simple-diff, token-down, tian2025bottlenecksampling, zhang2025scaledit}, 
{\color{red}
making it difficult to generate structurally complex objects with hundreds of parts.
}

Autoregressive generation offers an attractive alternative for addressing aforementioned limitations.
By modeling objects as structured token sequences, autoregressive models naturally align with part-aware object generation process. 
Moreover, recent advances in long-context large language models~\cite{megatron-lm,geminiteam2024gemini15unlockingmultimodal,bai2025qwen3vltechnicalreport} have substantially improved the practical feasibility of sequence modeling at unprecedented scales. 
Recent works~\cite{wang2024llamameshunifying3dmesh,ye2025shapellm,roblox2025cube,chen2024sar3d,deng2025efficientautoregressiveshapegeneration} have also shown the potential of autoregressive models in holistic object generation. 
However, directly applying autoregressive models to part-aware 3D generation is non-trivial. 
The core challenge is the tradeoff between geometric fidelity and token efficiency. 
Representing each part with a long token sequence can preserve fine geometric details, but the total sequence length grows rapidly with the number and complexity of parts, diminishing the scalability benefits of autoregressive models. 
On the other hand, reducing the token budget improves efficiency but often sacrifices the geometry fidelity of the generated shapes. 

In this work, we introduce \textbf{MegaParts}, an autoregressive framework for part-aware 3D object generation that scales to complex objects with hundreds of parts. 
To address the tradeoff challenge, we propose a token-efficient vector-quantized part tokenizer that learns an adaptive-length discrete representation for part-level geometry.
Specifically, the tokenizer encodes each part into a sequence of discrete latent tokens while encouraging geometric information to be concentrated in prefix tokens.
We then design a coding-rate distortion objective~\cite{shannon1959coding} to determine the token budget required for each part, yielding an efficient adaptive-length representation whose token length varies with geometric complexity. Our representation allows simple parts to be encoded with only a few tokens, while retaining sufficient capacity for geometrically intricate parts. Unlike conventional shape tokenizers~\cite{oord2018neuraldiscreterepresentationlearning, roblox2025cube} that focus primarily on reconstruction fidelity, our tokenizer jointly optimizes fidelity and token efficiency, making it well suited for large-scale autoregressive part-aware 3d generation of complex objects.

Building on this adaptive-length part representation, we train a large language model to generate 3D objects as unified structured sequences. Inspired by the chain-of-thoughts~\cite{wei2023chainofthoughtpromptingelicitsreasoning} nature of LLM reasoning, we represent the 3D object as a coherent assembly of localized parts. Specifically, the model first predicts the object-level bounding box, then generates the bounding box of each {\color{red} structural} part and finally produces the shape tokens within each predicted part bounding box. This design preserves explicit part structure throughout the generation process and supports fine-grained control at both object and part level.
Combined with efficient training strategies~\cite{megatron-lm}, our framework scales to objects with up to 300 parts and sequence lengths of up to 256k tokens, substantially expanding the practical scale of part-aware 3D generation compared to previous methods.

Importantly, this scalability is achieved without sacrificing generation quality. Experiments demonstrate that MegaParts generates higher-quality meshes compared to conventional autoregressive baselines~\cite{chen2024sar3d, roblox2025cube, ye2025shapellm} and even surpasses diffusion-based methods~\cite{xiang2024structured}, indicating that our adaptive-length discrete part representation improves both scalability and geometry fidelity. Moreover, these results suggest that LLM-native token-efficient autoregressive modeling is a compelling alternative to diffusion for large-scale part-aware 3D generation. 

In summary, our contributions are:
\begin{itemize}
    \item We propose a token-efficient vector-quantized tokenizer that learns adaptive-length discrete latent representations for part-level 3D geometry by minimizing token usage while maintaining high-fidelity reconstruction.
    \item We design a long-context autoregressive framework that generates object-level bounding box, part bounding boxes, and part geometry as a unified structured sequence.
    \item Our framework can scale to {\color{red} structurally} complex multi-part objects with up to 300 parts and 256k tokens, substantially extending the practical regime of part-aware 3D generation while improving geometric fidelity over prior baselines.
\end{itemize}

%% file: subsections/related_work.tex
\subsection{Representations for Structured 3D Shape Modeling}

Learning compact and expressive 3D shape representations has been widely studied across neural implicit fields~\cite{poole2022dreamfusion, lin2023magic3d, wang2023prolificdreamer}, VecSet representations~\cite{10.1145/3592442, zhao2023michelangelo, 10.1145/3658146, yang2024hunyuan3d, hunyuan3d22025tencent, 10.1145/3757377.3763812}, and voxel-based representations~\cite{xiang2024structured, xiang2025trellis2, wu2025direct3ds2gigascale3dgeneration}. VecSet representations~\cite{li2025craftsman3dhighfidelitymeshgeneration, direct3d, Chen_2025_Dora, li2025triposg} encode a shape using a shared pool of latent vectors to avoid the rigidity of regular grids. However, since global structure and local geometric details will compete for latent capacity, these representations often struggle to preserve small or intricate components in complex objects~\cite{lai2025latticedemocratizehighfidelity3d}.

Voxel-based representations~\cite{lai2025latticedemocratizehighfidelity3d, chen2025ultra3defficienthighfidelity3d, he2025triposf, lai2025hunyuan3d25highfidelity3d, luo2026topomeshhighfidelitymeshautoencoding} improve locality by assigning features to spatial regions. However, their capacity allocation is usually tied to a fixed grid or a predefined hierarchy, where geometrically simple and complex regions may receive similar representational budgets. Even adaptive variants~\cite{deng2025oat} still rely on manually designed spatial rules and do not naturally align with the semantic decomposition of component-rich or articulated objects.


These limitations motivate part-aware representations whose capacity can vary across parts according to geometric complexity. By separating global structure modeling from part-level representation, such representations are better suited for high-fidelity generation of complex shapes with hundreds of parts.
\vspace{-1mm}

\subsection{High-Fidelity Component-Aware 3D Generation Pipeline}

Existing part-aware 3D generation methods can be broadly divided into two categories. The first stream~\cite{yang2025holopart,tang2024partpacker,yan2025xparthighfidelitystructure,lin2025partcrafter,10.1145/3730840,chen2025autopartgenautogressive3dgeneration} assumes a part decomposition prior, such as segmentation masks or part IDs, and generates geometry conditioned on the prescribed partition. While such supervision provides explicit semantic guidance, it is often noisy, coarse, and insufficient for capturing internal structural details, which limits its applicability to generation of highly complex objects.

The second stream~\cite{ding2025fullpart,wang2025partxmllmpartaware3dmultimodal,dai2025meshcoder} adopts a native 3D generation formulation that jointly predicts part layouts and their corresponding geometries. This formulation is more expressive because it does not require a fixed part decomposition as input. However, existing methods in both streams are predominantly diffusion-based, which typically model both inter-part and intra-part relationships through carefully designed global and local attention mechanisms. This design restricts their scalability, with most methods focusing on objects containing fewer than 50 parts. As the number of parts increases, the cost of representing detailed geometry and maintaining coherent structural dependencies grows rapidly in token length, memory usage, and computational overhead, creating a fundamental bottleneck for complex part-aware 3D object generation.
\vspace{-1mm}

\subsection{Autoregressive Models for 3D Asset Modeling}

Autoregressive generation has become a popular paradigm for language and 2D visual targets driven by the success of GPT-style models. In contrast, autoregressive generation of 3D assets~\cite{chen2024sar3d, wei2025octgpt, Ibing_2023_CVPR, deng2025efficientautoregressiveshapegeneration, roblox2025cube, chen2024meshanything, lionar2025treemeshgpt, weng2024pivotmesh, weng2025scaling, xu2025meshmosaic, wang2024llamameshunifying3dmesh, hao2024meshtronhighfidelityartistlike3d, ICLR2025_58e6c003, zhao2025deepmesh, vonlutzow2026gaussiangptautoregressive3dgaussian} has only recently begun to attract substantial attention with most existing methods focus on directly generating vertices and faces for low-poly meshes~\cite{chen2024meshanything, lionar2025treemeshgpt, weng2024pivotmesh, weng2025scaling, xu2025meshmosaic, wang2024llamameshunifying3dmesh, hao2024meshtronhighfidelityartistlike3d, ICLR2025_58e6c003, zhao2025deepmesh}.

Beyond direct mesh tokenization, several methods~\cite{vonlutzow2026gaussiangptautoregressive3dgaussian,wei2025octgpt} explore structured discrete representations for autoregressive 3D generation. 
SAR3D~\cite{chen2024sar3d} constructs a coarse-to-fine representation based on multi-scale triplanes~\cite{Chan2021} and trains a VAR-like~\cite{NEURIPS2024_9a24e284} autoregressive model to generate this representation. 
Cube~\cite{roblox2025cube} offers a unified perspective on 3D generation and understanding by combining a simple vecset-based VQ-VAE with a pretrained GPT-2~\cite{radford2019language}. Octree Transformer~\cite{Ibing_2023_CVPR}, OctGPT~\cite{wei2025octgpt}, and OAT~\cite{deng2025efficientautoregressiveshapegeneration} adopt more structured octree-based representations and use autoregressive models to predict discrete hierarchical latents. 

Another line of work~\cite{ye2025shapellm, physxanything} adopts a two-stage pipeline, where an autoregressive model first generates a coarse structured geometry for global layout control, and a diffusion-based model~\cite{xiang2024structured} subsequently recovers fine geometry. ShapeLLM-Omni~\cite{ye2025shapellm} trains Qwen2.5-VL~\cite{bai2025qwen25vltechnicalreport} to generate coarse voxels and then uses the second stage of TRELLIS~\cite{xiang2024structured} to reconstruct meshes. 

Despite the growing popularity of autoregressive models for 3D generation, existing methods have not explored long-context autoregressive modeling for highly complex, part-aware 3D asset generation. This leaves open challenge of generating assets with many detailed components while preserving both global structural coherence and fine-grained part-level geometry.

%% file: subsections/3_1_vqvae.tex
\begin{figure*}[htp]
  \centering
  \includegraphics[width=\linewidth]{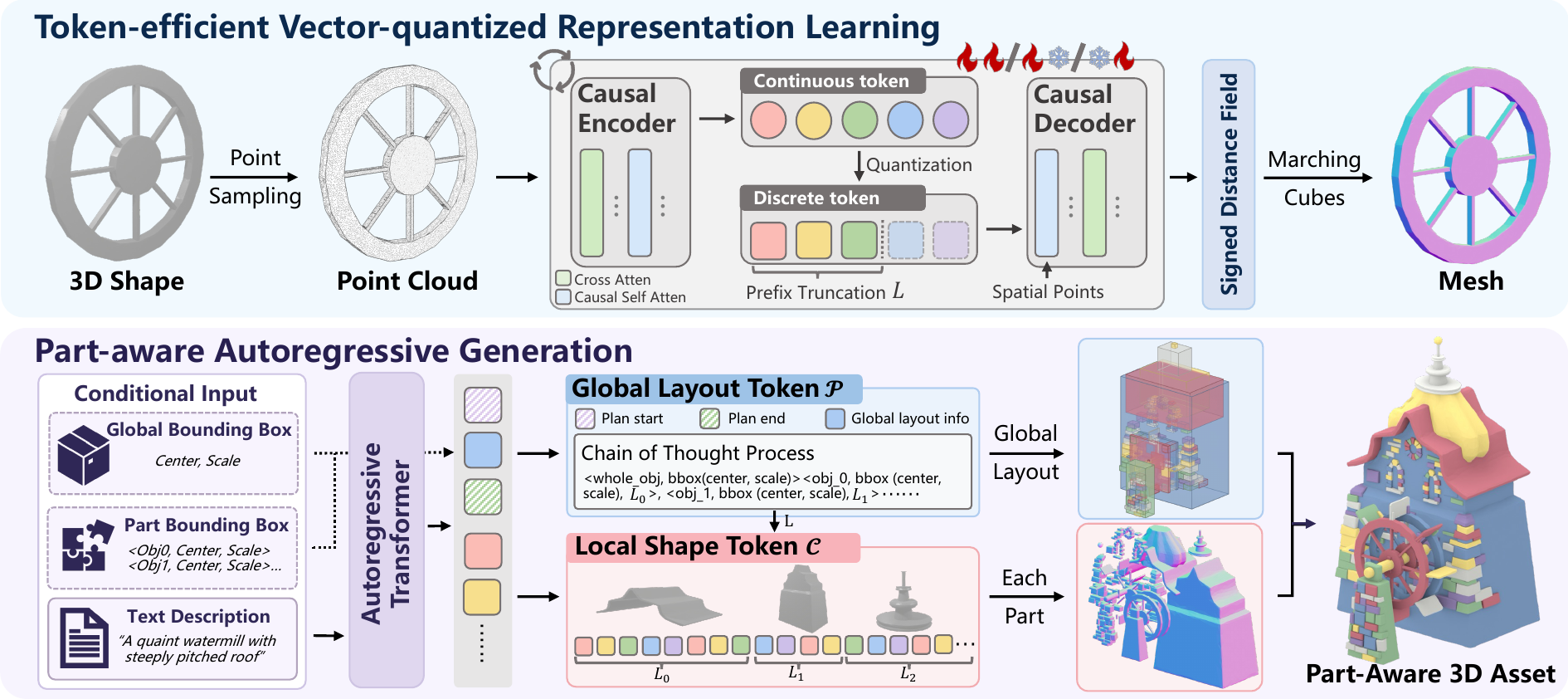}
  \caption{Overview of \textbf{MegaParts}. Our framework consists of two stages. Top: we learn a token-efficient vector-quantized representation for part geometry. Bottom: we train a part-aware autoregressive generator that represents a 3D object as a structured sequence conditioned on a text prompt and optional object- and part-level bounding boxes. If bounding boxes are absent, the model first predicts the object box and part boxes, followed by autoregressive generation of local shape tokens for each part. The decoded part geometries are then assembled into the final part-aware 3D asset.}
  \label{fig:method}
  \vspace{-2mm}
\end{figure*}

\subsection{Causal Autoencoder for Signed Distance Field Modeling}
\label{subsec:causal_1d_ae}

To support \emph{adaptive token budgets} for geometries with different levels of complexity, we introduce an adaptive-length discrete 3D shape representation based on a fully causal vector-quantized autoencoder. Under standard bidirectional attention, geometric information is often distributed relatively uniformly across the latent sequence, so truncating trailing tokens can significantly degrade the overall geometry. This behavior is undesirable for adaptive-length tokenization, where earlier tokens are expected to capture the most essential geometric content. To address this issue, we build our VQ-VAE upon the Cube~\cite{roblox2025cube} architecture, while replacing the bidirectional latent modeling in both the encoder and decoder with fully causal attention.

Given a surface point set \(\mathcal{S}=\{(\mathbf{p}_i,\mathbf{f}_i)\}_{i=1}^{N}\), where \(\mathbf{p}_i\in\mathbb{R}^3\) denotes the 3D coordinate of the \(i\)-th point and \(\mathbf{f}_i\) denotes its optional attribute, we first map the input into an embedded token sequence \(\mathbf{T}\in\mathbb{R}^{N\times d}\). The encoder maintains a learnable latent query array \(\mathbf{H}_0\in\mathbb{R}^{M\times d}\), where \(M\) is the latent sequence length and \(d\) is the feature dimension. At encoder layer \(\ell\), the latent queries are updated either through cross-attention to the input tokens or through causal self-attention:
\begin{equation}
\mathbf{H}_{\ell+1}=
\begin{cases}
\mathrm{CrossAttn}(\mathbf{H}_{\ell}, \mathbf{T}), & \ell\in\Omega_{\mathrm{cross}},\\
\mathrm{CausalSelfAttn}(\mathbf{H}_{\ell}), & \text{otherwise},
\end{cases}
\end{equation}
where \(\Omega_{\mathrm{cross}}\) denotes the set of encoder layers equipped with cross-attention. The final encoder output \(\mathbf{Z}_e\in\mathbb{R}^{M\times d}\) is then quantized by a spherical vector-quantization bottleneck~\cite{roblox2025cube} to obtain a discrete latent sequence \(\mathbf{Z}_q\in\mathbb{R}^{M\times d}\).

The decoder takes \(\mathbf{Z}_q\) together with learned positional embeddings and processes them using a stack of causal self-attention layers, yielding decoded latent features \(\mathbf{Y}\in\mathbb{R}^{M\times d}\). Because both the encoder and decoder are causal over the latent dimension, the representation naturally induces an information hierarchy in which earlier tokens are encouraged to encode more essential geometric content, making the model more robust to variable-length truncation. Finally, given a 3D query point \(\mathbf{x}\in\mathbb{R}^3\), a query head predicts its signed distance value by attending the query embedding to \(\mathbf{Y}\). The reconstructed mesh is then extracted from the predicted continuous signed distance field using marching cubes.

%% file: subsections/3_2_representation_training.tex
\subsection{Token-Efficient Stable VQ-VAE with Phased Optimization}
\label{subsec:token_efficient_stable_discrete_3d}


Building on the causal architecture above, we train the VQ-VAE with a stochastic truncation scheme to obtain a token-efficient and stable discrete representation. Specifically, during training we reconstruct shapes from random latent prefixes of length $L = \{2^n\}_{n=4}^{12}$, which encourages the encoder to concentrate geometric information into early tokens. Crucially, the causal decoder introduced in Section~\ref{subsec:causal_1d_ae} shields the informative prefix tokens from the noise of redundant trailing tokens, making variable-length tokenization possible without significant reconstruction degradation.

{
\color{red}
However, jointly optimizing the encoder and decoder through a discrete bottleneck often causes unstable code assignments, which can impede decoder convergence. To improve training stability, we go beyond the EMA-based stabilization strategy adopted in Cube~\cite{roblox2025cube} and introduce an alternating VQ-VAE optimization scheme. Following an initial stage of joint training, we alternate between encoder-only and decoder-only optimization phases, freezing the complementary module in each phase. Together with optimal-transport quantization~\cite{zhang2024preventing} and the Stochastic Gradient Shortcut (SGS)~\cite{roblox2025cube}, this training strategy stabilizes code assignments and promotes efficient codebook utilization. Combined with causal decoding, these techniques produce a robust and token-efficient discrete representation for 3D geometry.
}

We supervise the model with a weighted combination of reconstruction, eikonal regularization, surface normal alignment, and vector-quantization losses.
Let \(f_{\theta}(\mathbf{x};\mathbf{z})\) denote the predicted scalar field at query point \(\mathbf{x}\), conditioned on the latent sequence \(\mathbf{z}\). 
For a set of sampled query points \(\{\mathbf{x}_i\}_{i=1}^{N_1}\) with ground-truth signed distances \(\{s_i\}_{i=1}^{N_1}\), we use a truncated and normalized SDF (tsdf) target
\begin{equation}
\tilde{s}_i = \frac{\operatorname{clip}(s_i,-\delta,\delta)}{\delta},
\end{equation}
where \(\delta>0\) is the truncation threshold. 
The reconstruction term yields
\begin{equation}
\mathcal{L}_{\mathrm{tsdf}}
=
\frac{1}{N_1}\sum_{i=1}^{N_1}
\left(f_{\theta}(\mathbf{x}_i;\mathbf{z})-\tilde{s}_i\right)^2.
\end{equation}

To regularize the geometry of the predicted field, we impose an eikonal constraint on its spatial gradient. Let $\hat{s}_i = f_{\theta}(\mathbf{x}_i;\mathbf{z})$ and $\mathbf{g}_i = \nabla_{\mathbf{x}} \hat{s}_i.$
Rather than enforcing the constraint over the entire volume, we restrict it to a near-surface band
$\mathcal{B}=\left\{i \,\middle|\, |s_i| \le \alpha \delta \right\},$
where \(\alpha\) controls the band width. 
The eikonal loss is thus
\begin{equation}
\mathcal{L}_{\mathrm{eik}}
=
\frac{1}{|\mathcal{B}|}
\sum_{i\in\mathcal{B}}
\left(\|\mathbf{g}_i\|_2-1\right)^2.
\end{equation}

To further align the field gradient with surface orientation, we additionally sample surface points \(\{\mathbf{x}^{s}_j\}_{j=1}^{N_2}\) and their unit normals \(\{\mathbf{n}_j\}_{j=1}^{N_2}\). The predicted normal is obtained by normalizing the SDF gradient,
$
\hat{\mathbf{n}}_j
=
\nabla_{\mathbf{x}} \hat{s}(\mathbf{x}^{s}_j) /
\|\nabla_{\mathbf{x}} \hat{s}(\mathbf{x}^{s}_j)\|_2
$,
and the normal loss is defined by cosine alignment:
\begin{equation}
\mathcal{L}_{\mathrm{normal}}
=
\frac{1}{N_2}\sum_{j=1}^{N_2}
\left(1-\operatorname{clip}\!\left(\hat{\mathbf{n}}_j^{\top}\mathbf{n}_j,-1,1\right)\right).
\end{equation}

To align the encoder outputs with the codebook and update the codebook entries effectively, we adopt the standard vector-quantization loss~\cite{oord2018neuraldiscreterepresentationlearning}:
\begin{equation}
\mathcal{L}_{\mathrm{vq}}
=\left\|\mathbf{Z}_e-\mathrm{sg}[\mathbf{Z}_q]\right\|_2^2
+\left\|\mathbf{Z}_q-\mathrm{sg}[\mathbf{Z}_e]\right\|_2^2,
\end{equation}
where $\mathrm{sg}[\cdot]$ denotes the stop-gradient operator.

The full training objective is
\begin{equation}
\mathcal{L}
=
\lambda_{\mathrm{tsdf}}\mathcal{L}_{\mathrm{tsdf}}
+
\lambda_{\mathrm{eik}}\mathcal{L}_{\mathrm{eik}}
+
\lambda_{\mathrm{normal}}\mathcal{L}_{\mathrm{normal}}
+
\lambda_{\mathrm{vq}}\mathcal{L}_{\mathrm{vq}},
\end{equation}
where \(\lambda_{\mathrm{tsdf}}\), \(\lambda_{\mathrm{eik}}\), \(\lambda_{\mathrm{normal}}\) and \(\lambda_{\mathrm{vq}}\)  are scalar hyperparameters controlling the relative importance of these four terms.

%% file: subsections/3_3_token_count.tex
\subsection{Adaptive-Length Representation via Rate--Distortion Optimization}
\label{subsec:token_rep_rate_dis}

\begin{figure}[!b]
  \centering
  \includegraphics[width= 0.85\linewidth]{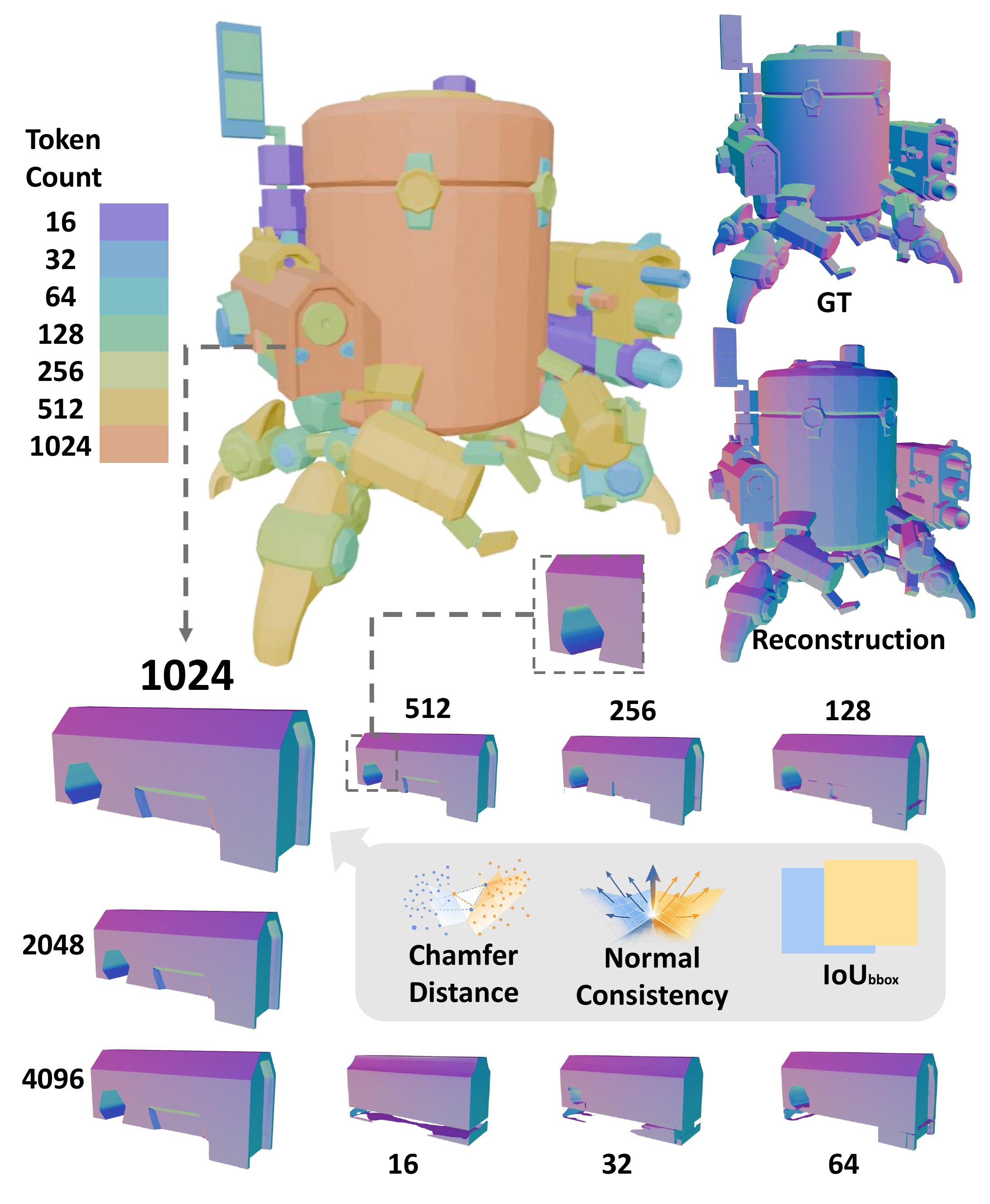}
  \caption{Given a complex object consisting of multiple parts, we adaptively assign each part a suitable token budget. During inference of the VQ-VAE we reconstruct part meshes under different token budgets and determine the optimal token number based on a carefully designed rate distortion metric. Our method enables flexible representation {\color{red}volume} for part meshes with different complexity while introducing minimum visual degradation.}
\end{figure}


Given the specially trained VQ-VAE from Section~\ref{subsec:token_efficient_stable_discrete_3d}, we estimate an appropriate token length for each shape using a criterion motivated by rate--distortion theory~\cite{shannon1959coding}. To enable adaptive representation for complex objects, we treat each component as an independent encoding unit. Specifically, a global mesh is decomposed as $\mathcal{G} = \{\mathcal{M}^{(c)}\}_{c=1}^{K}$, where each $\mathcal{M}^{(c)}$ denotes a semantic or structural part (e.g., a table leg) and ${K}$ is the number of parts in the global mesh. 
The VQ-VAE encodes each component in its local $[-1,1]$ normalized coordinate frame, while its bounding box is recorded separately to recover its global position and scale.

For component \(c\), the encoder produces a full discrete latent sequence $\mathbf{z}^{(c)}=(z^{(c)}_1,\ldots,z^{(c)}_{L_{\max}})$.
Consistent with the prefix truncation strategy used during training, we evaluate each candidate prefix length $L$
and decode only the prefix \(\mathbf{z}^{(c)}_{1:L}\), producing a reconstructed component mesh $\hat{\mathcal{M}}^{(c)}_{L}.$

The optimal representation length for component \(c\) is selected as
\begin{equation}
L_c^\star
=
\mathop{\arg\,\min}\limits_{L\in \left\{2^n\right\}_{n=4}^{12}} J^{(c)}(L),
\end{equation}
where
{\color{red}
\begin{equation}
J^{(c)}(L)
=
D^{(c)}(L)
+
\gamma R^{(c)}(L),
\end{equation}
}
is the rate--distortion objective~\cite{shannon1959coding}. Here,
\begin{equation}
D^{(c)}(L)
=
w_{\mathrm{cd}} D_{\mathrm{cd}}^{(c)}(L)
+
w_{\mathrm{n}} D_{\mathrm{normal}}^{(c)}(L)
+
w_{\mathrm{u}} D_{\mathrm{bbox}}^{(c)}(L),
\end{equation}
where \(D_{\mathrm{cd}}^{(c)}(L)\) is the Chamfer distance between the reconstructed mesh and the reference mesh, \(D_{\mathrm{normal}}^{(c)}(L)\) measures their surface normal inconsistency and \(D_{\mathrm{bbox}}^{(c)}(L)\) measures the IoU between the meshes to avoid small floaters. Together, these terms encourage faithful geometric reconstruction. The rate term $R^{(c)}(L)=L/L_{\max}$ penalizes the number of tokens used to represent the component and {\color{red} $\gamma$} is the penalty strength.

The resulting representation of the full object is
$\mathcal{Z}_{\mathcal{G}}
=
\{
(
\mathbf{z}^{(c)}_{1:L_c^\star},
\mathbf{b}^{(c)}
)
\}_{c=1}^{K},$
where \(\mathbf{b}^{(c)}\) is the bounding box of component \(c\) in the global coordinate frame. This representation stores geometry in normalized component codes, while the bounding boxes recover the layout and scale of all parts in the original complex shape.

%% file: subsections/3_4_autoregressive.tex
\subsection{Component Aware Generation With Structured Autoregressive Modeling}
\label{subsec:com_gui_str_rep}

Given the component-level adaptive representation above, we further build a structured global shape representation inspired by the reasoning process of large language models~\cite{wei2023chainofthoughtpromptingelicitsreasoning}. 
Since the VQ-VAE is trained on component meshes normalized to a centered \([-1,1]\) canonical space, each generated component should be placed back into the global coordinate system. 
We therefore factorize the shape sequence generation into two stages: layout generation and geometry generation.

For a mesh with \(K\) components, we represent the global layout as
$\mathcal{P}
=
[
\mathbf{a}^{(1)},\mathbf{a}^{(2)},\ldots,\mathbf{a}^{(K)}
],$
where each component descriptor is
$\mathbf{a}^{(c)}
=
[
\mathbf{b}^{(c)},L_{c}^*
].$
Here, $\mathbf{b}^{(c)}=[\mathbf{m}^{(c)},\mathbf{e}^{(c)}]$, \(\mathbf{m}^{(c)}\in\mathbb{R}^{3}\) is the component bounding-box center, \(\mathbf{e}^{(c)}\in\mathbb{R}^{3}\) is the bounding-box scale. The corresponding component geometry is represented by the discrete mesh-code sequence
$\mathcal{C}
=
[
\mathbf{z}^{(1)}_{1:L_{1}^*},
\mathbf{z}^{(2)}_{1:L_{2}^*},
\ldots,
\mathbf{z}^{(K)}_{1:L_{C}^*}
].$
The full global representation is thus
$\mathcal{Y}
=
\left[
\mathcal{P},\mathcal{C}
\right],$
which corresponds to first generating the layouts of all components and then generating the geometry tokens of each component autoregressively. This decomposition makes the representation explicit, interpretable, and naturally suitable for part-level control.

To improve sequence regularity, we impose a deterministic ordering over components. At the global level, components are sorted according to the spatial order of their bounding-box centers, which encourages the model to learn consistent structural dependencies among nearby parts. At the local level, each component is represented by a prefix-based token sequence, progressing from coarse geometric to finer details. Together, this representation combines a stable global layout with token-efficient local geometry encoding.





Building on this structured representation, we formulate 3D generation as autoregressive sequence modeling. We fine-tune a pretrained large language model, \textsc{Qwen3}~\cite{qwen3} as the generator. The model is trained with a standard next-token prediction objective:
$p_{\theta}(\mathbf{s}\mid \mathbf{q})
=
\prod_{t=1}^{T}
p_{\theta}(s_t \mid \mathbf{q}, s_{<t}),$
where \(\mathbf{q}\) denotes the prompt containing control signals and \(\mathbf{s}=(s_1,\ldots,s_T)\) is the serialized shape representation. To efficiently handle the resulting long, structured shape sequences, we fine-tune the LLM using context and tensor parallelism~\cite{megatron-lm}, ensuring both training efficiency and scalability. The object and part bounding box tokens are jointly optimized with the shape tokens. This ensures that the model can perform both box conditional and unconditional training.

To reconstruct a valid mesh from the generated sequence, we first decode the predicted part bounding box tokens into bounding boxes $\hat{\mathbf{b}}^{(c)}=[\hat{\mathbf{m}}^{(c)},\hat{\mathbf{e}}^{(c)}]$. We then apply the VQ-VAE decoder to the generated part shape tokens and obtain the part mesh $\hat{\mathcal{M}}^{(c)}$ via marching cubes~\cite{10.1145/37401.37422} in canonical space. Let $\mathbf{v}_{\hat{\mathcal M}^{(c)}}$ denote the vertices of $\hat{\mathcal{M}}^{(c)}$. The part is mapped back to the global coordinate system by {\color{red} isotropic} scaling and translation, i.e., $0.5\max(\hat{\mathbf{e}}^{(c)})\cdot \mathbf{v}_{\hat{\mathcal M}^{(c)}} + \hat{\mathbf{m}}^{(c)}$. The full object mesh is obtained by assembling all transformed part meshes.
\vspace{-1mm}

%% file: subsections/6_1_experiment_intro.tex

\begin{table}[t]
\centering
\resizebox{\linewidth}{!}
{
\setlength{\tabcolsep}{10pt}
\begin{tabular}{lcccc}
\toprule
& \multicolumn{2}{c}{Part-level} & \multicolumn{2}{c}{Object-level} \\
\cmidrule(lr){2-3} \cmidrule(lr){4-5}
Method
& CD($\times 10^{-3}$) $\downarrow$ & NC $\uparrow$
& CD($\times 10^{-3}$) $\downarrow$ & NC $\uparrow$ \\
\midrule
Cube (1024) & 3.95 & 0.89 & 1.89 & 0.85 \\
Ours (1024) & 0.12 & 0.93 & 1.52 & 0.89 \\
\midrule
Ours (512)  & 0.35 & 0.90 & 1.92 & 0.80 \\
Ours (2048) & 0.07 & 0.97 & 1.32 & 0.93 \\
Ours (4096) & 0.06 & 0.98 & 1.25 & 0.94 \\
\bottomrule
\end{tabular}
}
\caption{Quantitative reconstruction results of VQ-VAEs at both the part and object levels. At the matched budget of 1024 tokens, our model consistently outperforms Cube. Moreover, increasing the token budget from 512 to 4096 yields progressively better reconstruction quality, indicating a clear coarse-to-fine trend in the learned discrete representation.}
\label{tab:exp_recon_partobjtiny}
\vspace{-6mm}
\end{table}

\subsection{Evaluation Protocol}
\label{subsec:exp_intro}


Our pipeline is trained on a curated dataset, including public and private data, as described in Supplementary Material A2. For fair comparison, we evaluate our method on PartObjaverse-Tiny~\cite{yang2024sampart3d}, a publicly available benchmark for part-aware 3D modeling. {\color{red} Additional results on high-part-count object evaluation can be seen in Supplementary Material C1.} We assess both the reconstruction and generative capabilities of the proposed framework. 
\vspace{-3mm}

\begin{table}[tp]
\centering
\resizebox{\linewidth}{!}{
\begin{tabular}{lccccc}
\hline
 & SAR3D & Cube & TRELLIS-text & ShapeLLM-Omni & Ours \\
\hline
FID $\downarrow$  & 94.69 & 55.58 & 54.81 & 70.99 & \textbf{43.40} \\
CLIP $\uparrow$   & 0.21 & 0.26 & 0.25 & 0.23 & \textbf{0.27} \\
\hline
\end{tabular}
}
\caption{Generation evaluation on PartObjaverse-Tiny with normal-map--based semantic alignment. Our method achieves better performance over text-conditioned generation baselines.}
\label{tab:text_gen_results}
\vspace{-7mm}
\end{table}

\begin{table}[tp]
\centering
\resizebox{\linewidth}{!}{
\setlength{\tabcolsep}{18pt}
\begin{tabular}{lccc}
\hline
Metric & FullPart & XPart & Ours \\
\hline
Part CD($\times 10^{-2}$) $\downarrow$ & 8.59 & 8.01 & \textbf{3.01} \\
Part IoU $\uparrow$ & 0.41 & 0.52 & \textbf{0.63} \\
BBox IoU $\uparrow$ & 0.76 & 0.59 & \textbf{0.94} \\
\hline
\end{tabular}
}
\caption{Results for part-bounding-box-conditioned shape generation. Our method achieves the best performance among the compared baselines.}
\label{tab:bbox_gen_results}
\vspace{-8mm}
\end{table}

\subsection{Evaluation of VQ-VAE}

For VQ-VAE reconstruction, we report Chamfer Distance (CD) and Normal Consistency (NC). 
Before evaluation, all meshes are normalized to a centered canonical space of \([-1,1]^3\), and 100{,}000 points are sampled from each mesh to compute both metrics. 
We use Cube~\cite{roblox2025cube} as the primary baseline, as it is the open-source VQ-VAE most closely matched to our setting. 

To ensure fairness, our main comparison uses the token counts adopted by the baseline during training. 
Since our model additionally supports adaptive reconstruction under varying token budgets, we further evaluate it across multiple token-count settings. We present reconstruction metrics in Table~\ref{tab:exp_recon_partobjtiny}.


\begin{table*}[t]
\centering
\small
\setlength{\tabcolsep}{11pt}
\resizebox{0.98\linewidth}{!}{
\begin{tabular}{ll cc cc cc cc}
\toprule
Setting & Scope
  & \multicolumn{2}{c}{512} & \multicolumn{2}{c}{1024}
  & \multicolumn{2}{c}{2048} & \multicolumn{2}{c}{4096} \\
\cmidrule(lr){3-4} \cmidrule(lr){5-6} \cmidrule(lr){7-8} \cmidrule(lr){9-10}
& & CD($\times 10^{-3}$) $\downarrow$ & NC $\uparrow$ & CD($\times 10^{-3}$) $\downarrow$ & NC $\uparrow$
  & CD($\times 10^{-3}$) $\downarrow$ & NC $\uparrow$ & CD($\times 10^{-3}$) $\downarrow$ & NC $\uparrow$ \\
\midrule
\multirow{2}{*}{Ours (full)}
  & Part & \textbf{0.35} & \textbf{0.90} & \textbf{0.12} & \textbf{0.93} & \textbf{0.07} & \textbf{0.97} & \textbf{0.06} & \textbf{0.98}  \\
  & Object & \textbf{1.92} & \textbf{0.80} & \textbf{1.52} & \textbf{0.89} & \textbf{1.32} & \textbf{0.93} & \textbf{1.25} & \textbf{0.94} \\
\midrule
\multirow{2}{*}{w/o phased {\color{red}VQ-VAE} training}
  & Part & 0.42 & 0.77 & 0.22 & 0.82 & 0.18 & 0.83 & 0.18 & 0.88  \\
  & Object & 3.05 & 0.76 & 2.88 & 0.81 & 2.51  & 0.82 & 2.17 & 0.87 \\
\midrule
\multirow{2}{*}{w/o causal attn.}
  & Part & 0.66 & 0.89 & 0.22 & 0.90 & 0.13 & 0.93 & 0.10 & 0.95  \\
  & Object & 2.77 & 0.82 & 1.92 & 0.86 & 1.51 & 0.92 & 1.33 & 0.92 \\
\bottomrule
\end{tabular}
}
\caption{Ablation study of phased training and causal attention in our VQ-VAE. We report part-level and object-level reconstruction metrics under different token budgets. The full model performs best across all settings, while removing phased training degrades reconstruction quality and removing causal decoder attention weakens the benefit of larger token budgets.}
\label{tab:vqvae_ablation_part_obj_tokens_wide}
\vspace{-4mm}
\end{table*}

\subsection{Evaluation of the Generation Model}
We evaluate generative performance on two tasks: \emph{text-conditioned 3D generation} and \emph{part-bounding-box-conditioned shape generation}. For text-conditioned generation, we compare against SAR3D~\cite{chen2024sar3d}, {\color{red} Cube~\cite{roblox2025cube},} TRELLIS-text~\cite{xiang2024structured}, and ShapeLLM-Omni~\cite{ye2025shapellm}.


{\color{red} We additionally introduce \emph{part-bounding-box-conditioned shape generation} to evaluate structure-aware synthesis. This task conditions generation on layout specified by part bounding boxes together with auxiliary inputs. We compare against FullPart~\cite{ding2025fullpart} and XPart~\cite{yan2025xparthighfidelitystructure} on this setting.}

Following prior work~\cite{xiang2024structured, ye2025shapellm, chen2024sar3d}, we evaluate text-conditioned generation using FID and CLIP score computed on rendered normal maps, and further present qualitative comparisons in Figure~\ref{fig:text_cond}. For part-bounding-box-conditioned generation, we report part-level IoU, part-level Chamfer Distance and BBox IoU. The quantitative results are summarized in Table~\ref{tab:text_gen_results} and Table~\ref{tab:bbox_gen_results}. {\color{red}
For a controlled comparison, we select baselines with an explicit per-part bounding-box prediction stage~\cite{ding2025fullpart,yan2025xparthighfidelitystructure} and replace predicted boxes with ground-truth part bounding boxes. This ensures that all methods receive the same structural layout, thereby controlling differences introduced by original input modalities.
}


\paragraph{Discussion.} The baseline methods often exhibit a failure mode in which multiple semantic or structural parts are generated as a single fused mesh, rather than as distinct components like that in Figure~\ref{fig:part_cond}. Because our evaluation is performed at the part level, such fused outputs are heavily penalized during normalization and metric computation, which can lead to disproportionately poor Chamfer Distance or IoU values. 
{\color{red} Therefore, these metrics primarily measure adherence to the prescribed part decomposition and layout, rather than overall appearance quality across different conditioning modalities. The results in Table~\ref{tab:bbox_gen_results} should be interpreted together with the qualitative comparisons, rather than as a fully controlled cross-modality appearance benchmark.}

To further assess the capability of our model, we additionally compare with image-conditioned part-aware generation methods~\cite{10.1145/3757377.3763872, tang2024partpacker, lin2025partcrafter}. Because these methods rely on a different input modality, there is no straightforward and fully fair numerical protocol for direct comparison. We therefore present representative qualitative comparisons in Figure~\ref{fig:part_aware}.



%% file: subsections/7_1_ablation_intro.tex
\subsection{Ablation Study}
We ablate two core design choices in our 3D VQ-VAE: causal attention and {\color{red} alternating VQ-VAE optimization} in Table~\ref{tab:vqvae_ablation_part_obj_tokens_wide} using PartObjaverse-Tiny. Dropping phased {\color{red}VQ-VAE} optimization consistently hurts reconstruction metrics, indicating that {\color{red}staged VQ-VAE optimization} stabilizes optimization over discrete latents. Removing causal attention causes a similarly clear drop in performance, which supports using a causal decoder for effective reconstruction when token budgets vary adaptively. {\color{red}Ablations on the generation model are presented in Supplementary Material B.}

%% file: subsections/8_1_detail_second_time_creation.tex
\begin{figure}[!b]
  \centering
  \includegraphics[width=\linewidth]{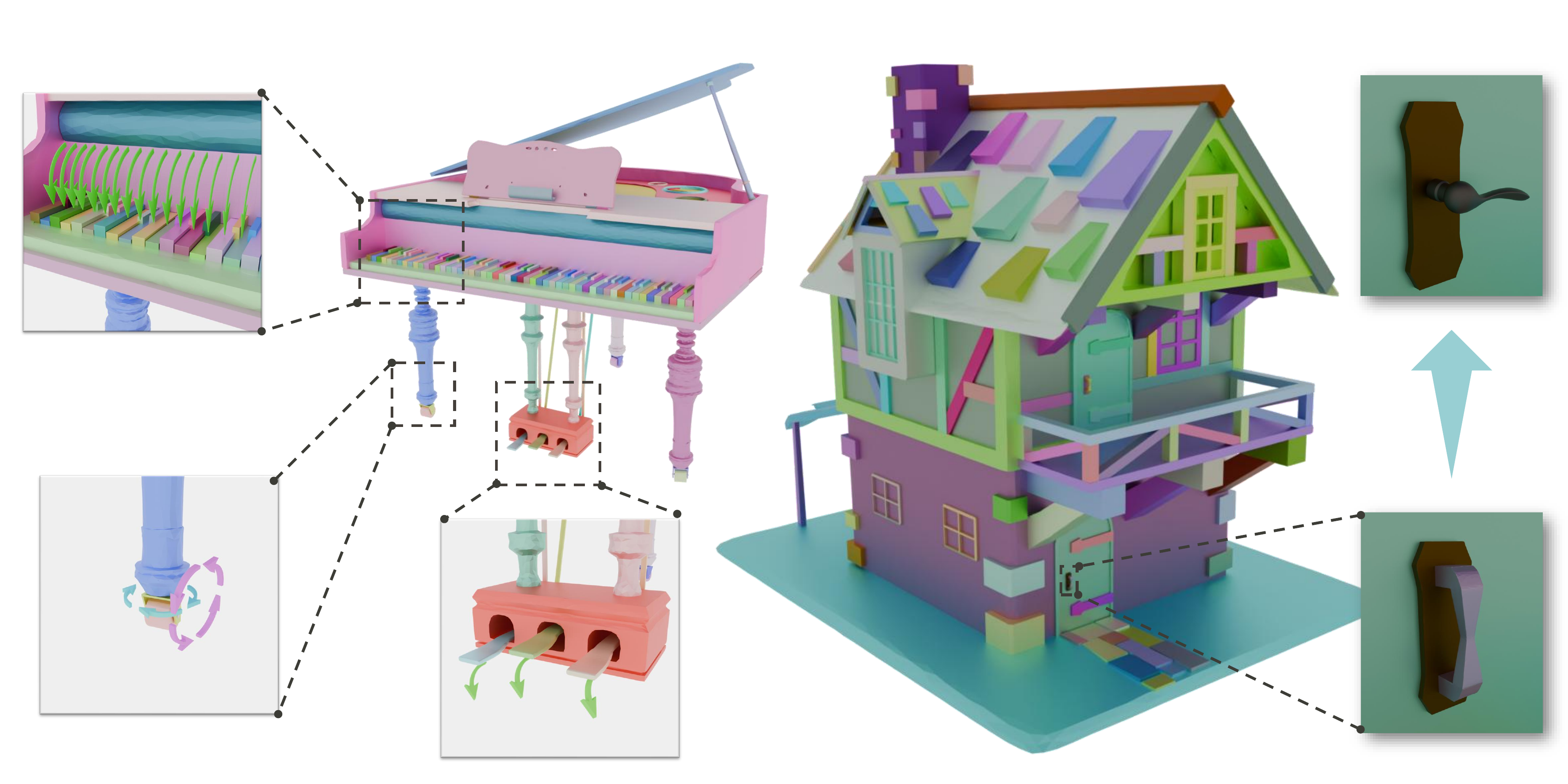}
  \caption{Fine-grained part-aware generation supports diverse downstream applications. Left: articulated object creation with detailed part decomposition. Right: secondary creation guided by localized part priors.}
  \label{fig:application}
  \vspace{-6mm}
\end{figure}

Figure~\ref{fig:application} shows two representative downstream applications of our generated results. With detailed part decomposition, our framework enables fine-grained secondary creation and articulation assets creation.

\subsection{Part-Level Editing and Secondary Creation}

In addition to improving generation quality, the fine-grained part structure produced by our model is directly useful for digital content creation. Each generated asset is decomposed into detailed components, providing an explicit part-level prior for localized manipulation. This allows artists to edit specific regions, recombine parts from different assets, and refine individual components without reconstructing the entire shape. Such structured outputs support flexible secondary creation, narrowing the gap between automatic 3D generation and practical artist-driven workflows.

%% file: subsections/8_2_application_articultation_creation.tex
\subsection{Fine-Grained Articulated Assets Creation}
The proposed fine-grained segmentation prior is particularly useful for generating articulated assets with dense interactive structures. Unlike monolithic representations, our part-aware generation explicitly separates small functional components, such as the individual keys of a piano or the buttons and switches of a control panel. This detailed decomposition allows each component to be associated with its own motion parameters, interaction logic, or physical constraints, thereby facilitating the construction of interactive and animatable 3D assets. As a result, the generated models can serve not only as static geometry, but also as structured assets for animation, simulation, and digital interaction.

%% file: subsections/9_1_c_d.tex
{\color{red}We present MegaParts, a scalable autoregressive framework for fine-grained part-aware 3D generation. Empowered by a token-efficient vector-quantized tokenizer with adaptive-length token selection, and a long-context autoregressive model, MegaParts generates objects with up to 300 parts while preserving geometric fidelity and structural coherence. It outperforms prior autoregressive baselines and remains competitive with diffusion-based methods.

Despite the impressive experimental results of our method, several limitations remain. Long-context autoregressive generation still incurs substantial training and inference cost, especially for highly complex objects. The current framework models geometry only, without textures, materials, semantic annotations, or physical and functional relationships among parts. Future work will explore more efficient sequence modeling, hierarchical representations, and unified modeling of geometry, appearance, and semantics.
}

%% file: subsections/a_1_1_model_sepc.tex
\subsection{Implementation Details of Models}
\label{subsec:model_spec}

Our pipeline consists of two key components: an adaptive 3D VQ-VAE for part-level shape representation and a fine-tuned large language model (LLM) for conditional generation.

\paragraph{Adaptive 3D VQ-VAE}
The architecture of our VQ-VAE is described in Section 3.1. Our model follows a structure similar to Cube v0.5~\cite{roblox2025cube}, which allows us to initialize it from the pretrained Cube weights. However, Cube v0.5 uses only 1024 learnable shape-query tokens and a codebook with 16,384 quantization entries. To increase the representation capacity, we expand both the query tokens and the codebook by copying the original weights three times and adding small random perturbations to the copied entries to encourage diversity. This yields a model with 4096 query tokens and a 65,536-entry codebook, which better supports the token-efficient and stable discrete 3D representation described in Section 3.2.

\begin{figure}[h]
  \centering
  \includegraphics[width=\linewidth]{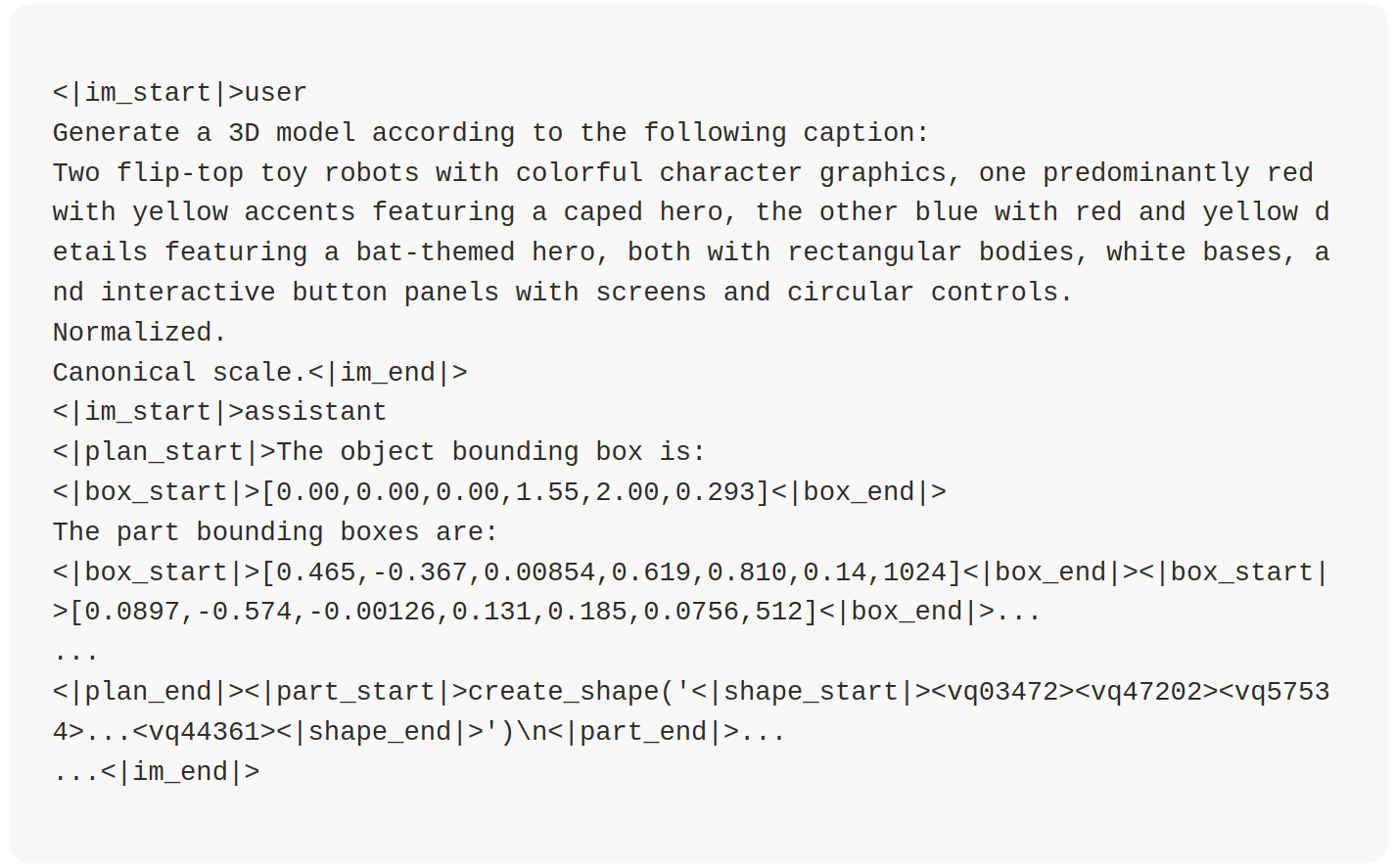}
  \caption{Example of the text processed by \textsc{Qwen3}.}
  \label{fig:qwen_input}
\end{figure}

\paragraph{Autoregressive LLM}
We build our generator on a pretrained \textsc{Qwen3}-8B~\cite{qwen3} language model and extend its vocabulary with task-specific structural and geometry tokens. We introduce a dedicated set of discrete shape tokens,
\(\{\texttt{<vq00000>}, \ldots, \texttt{<vq65535>}\}\),
to represent the codebook entries produced by the VQ-VAE tokenizer. To mark the boundaries of geometry token spans, we additionally introduce \texttt{<|shape\_start|>} and \texttt{<|shape\_end|>}, which denote the beginning and end of a sequence of shape tokens for a component. At the part level, \texttt{<|part\_start|>} and \texttt{<|part\_end|>} are used to delimit each component record. To explicitly separate structure planning from geometry generation, we further introduce \texttt{<|plan\_start|>} and \texttt{<|plan\_end|>}, which enclose the planning sequence containing the object bounding box and all part bounding boxes. The object and part bounding boxes are encoded by float numbers with three effective digits, which are encapsulated by special tokens \texttt{<|box\_start|>} and \texttt{<|box\_end|>}. These special tokens allow the model to represent global object layout, part structure, and part geometry within a unified serialized sequence while preserving clear boundaries between different semantic fields. Figure~\ref{fig:qwen_input} provides an example of the resulting text sequence processed by the LLM.


%% file: subsections/a_1_2_data_collection_preprocess.tex
\subsection{Data Collection and Processing}

Data pipelines play a critical role in training high-quality 3D representations and large-scale 3D generation models. We curate nearly 10M part meshes for VQ-VAE training, together with an elaborately processed, well-annotated, part-aware 3D mesh dataset of approximately 440K assets for generation model training. Across the full training pipeline, we use PartVerse-XL~\cite{ding2025fullpart}, PartNext~\cite{NEURIPS2025_579636b6}, PartNet~\cite{chen2025autopartgenautogressive3dgeneration}, Infinigen~\cite{infinigen2023infinite,infinigen2024indoors}, Infinite-Mobility~\cite{lian2025infinitemobilityscalablehighfidelity}, HSSD~\cite{khanna2023hssd}, HY3D-Bench~\cite{hunyuan3d2026hy3dbenchgeneration3dassets}, ShapeNet~\cite{chang2015shapenetinformationrich3dmodel}, Objaverse~\cite{objaverse}, and a high-quality private dataset. {\color{red} Among them, the public dataset contains 340k assets and the private dataset contains 100k assets.}

\begin{figure*}[htp]
  \centering
  \includegraphics[width=0.8\linewidth]{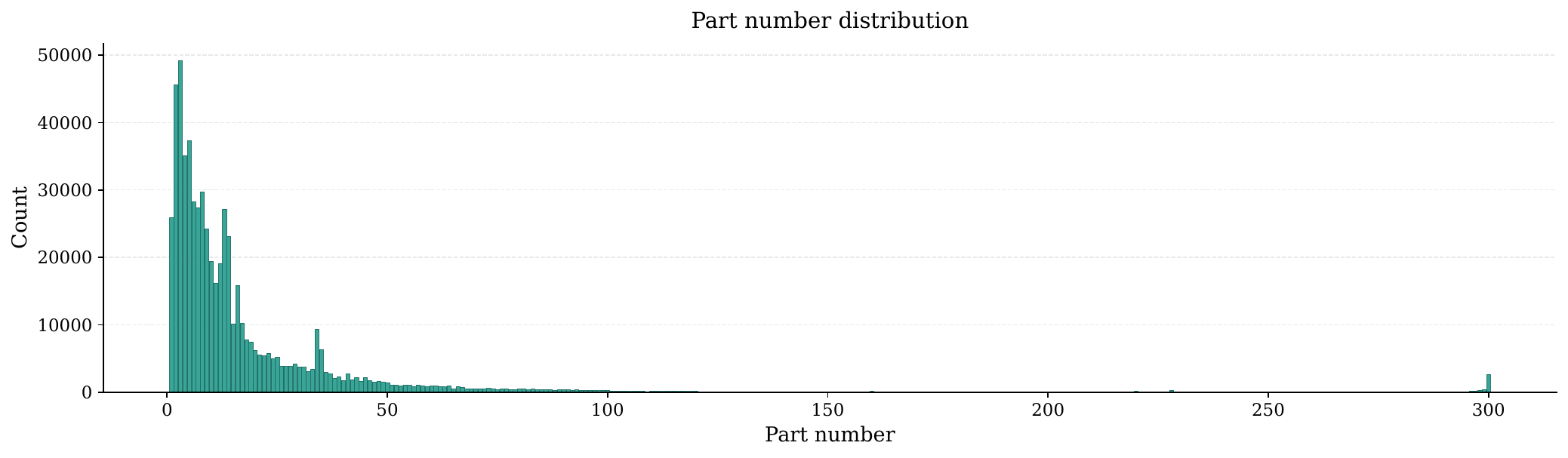}
  \caption{Distribution of the number of parts for meshes in the dataset.}
  \label{fig:appendix_dist_part}
\end{figure*}

\begin{figure*}[htp]
  \centering
  \includegraphics[width=0.8\linewidth]{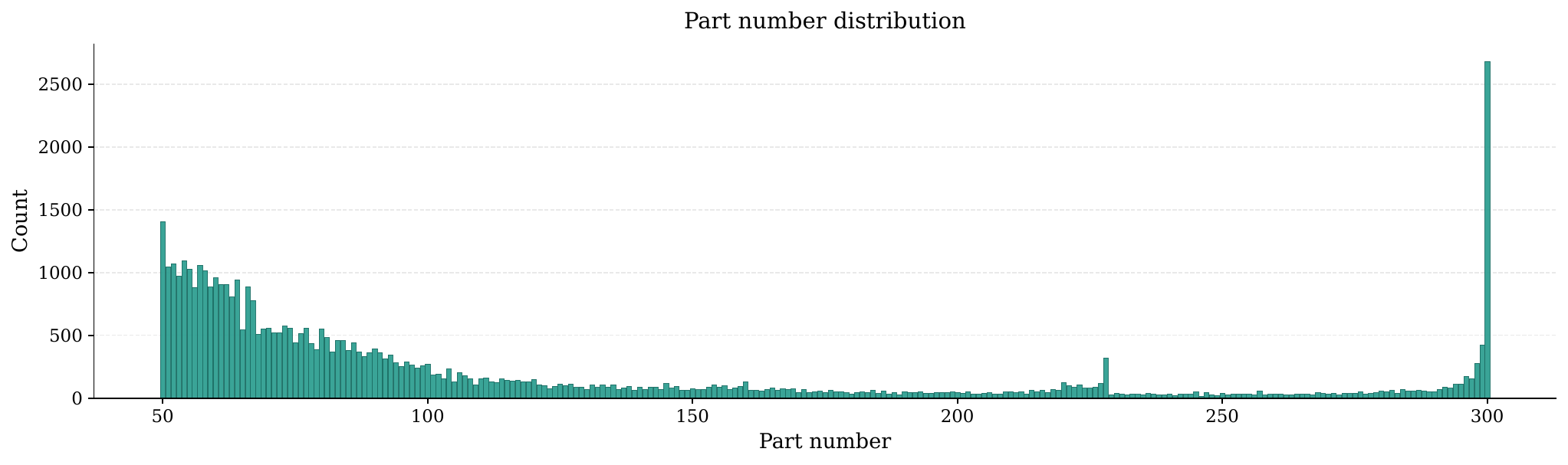}
  \caption{Distribution of the number of parts for meshes with more than 50 parts in the dataset.}
  \label{fig:appendix_dist_part_50}
\end{figure*}

\paragraph{Mesh Preprocessing.}
Human-annotated part-aware 3D datasets remain scarce. Following the practice of Hy3D-Bench, we therefore decompose object-level meshes from diverse sources into connected components to obtain more diverse part-level training data. For meshes with more than 300 connected components, we merge components according to their bounding-box sizes and relative spatial positions. {\color{red} Specifically, we rank parts according to the bounding box sizes and build a connectivity graph using the voxelized mesh. We then iteratively merge the smallest component into the adjacent component with which it has the largest voxel overlap, until at most 300 parts remain. This procedure yields a part decomposed dataset of up to 300 parts. The distribution of the part number is shown in Figure~\ref{fig:appendix_dist_part}. We also visualize the distribution of assets with more than 50 parts separately, as shown in Figure~\ref{fig:appendix_dist_part_50}.}

{\color{red}However, parts obtained through connected-component decomposition are often non-manifold and non-watertight, which makes both SDF construction and geometry reconstruction difficult.} To address this issue, we develop a multi-stage hole-filling pipeline. For each original mesh, we sequentially apply three repair strategies: Blender's 3D Print Toolbox~\cite{print3dtoolbox2026} for manifold mesh cleaning, VolumeMaker~\cite{volmesh2021}, and Blender's solidify modifier. After each attempt, we compute the Chamfer distance between the repaired mesh and the original mesh. We select the first repaired result whose distance falls below a desired threshold. After hole filling, we use cubvh~\cite{cubvh2026} for mesh watertightening and ground-truth SDF value extraction.

\paragraph{Data Recipes.}
For VQ-VAE training, our goal is to learn a robust representation that can handle inputs with arbitrary geometric complexity, including object-level complexity. Although the VQ-VAE is mainly used for part-level inference in our pipeline, we train it with a mixture of object-level meshes and part-level meshes. Empirically, we observe that when the proportion of object-level meshes is too small, reconstructed shapes tend to lose details and contain holes. We therefore adopt a 1:1 mixture of object-level and part-level meshes to improve reconstruction quality and robustness.

For object-level part-aware generation training, we filter out samples whose token sequence length exceeds 256k, as well as samples containing overly complex part geometries for which the VQ-VAE cannot produce valid reconstructions. This filtering improves training stability while preserving high-quality assets with detailed component structures. {\color{red} To mitigate imbalance across objects of different number of parts}, we further rebalance the training data according to part count. Specifically, we divide the samples into four groups with part counts in the ranges $[1,50]$, $(50, 100]$, $(100, 200]$ and $(200, 300]$, and sample from these groups with a uniform mixing ratio of $1:1:1:1$ during training.

\paragraph{Data Annotations.}
The success of text-controlled 3D asset generation heavily relies on high-quality annotation captions. Thus, we carefully design a data annotation pipeline to generate accurate and detailed captions. To capture comprehensive geometry, following Trellis~\cite{xiang2024structured}, we render 8 views from spherical camera poses. The yaw angles are sampled to approximately uniformly cover the full 360° range, while the pitch angles gradually increase from low to high latitudes, ensuring coverage of both the equator and upper regions. A small random perturbation is applied to the sampling order for each 3D asset, introducing some randomness into the view distribution. This yields a set of images that jointly cover most of the mesh surface with systematic yet diverse sampling. We then employ one of the leading multimodal models, Kimi-K2.5~\cite{kimiteam2026kimik25visualagentic}, to generate captions for each data sample. Specifically, we ask Kimi-K2.5 to produce the following information to support effective and robust training: (1) hierarchical category information; (2) mesh condition; (3) material information; (4) ten textual descriptions whose granularity varies from fine to coarse. We subsequently apply a filtering and label-balancing strategy, removing data samples with incomplete or damaged geometry to ensure valid mesh structure.

%% file: subsections/a_1_3_training_recipes.tex
{
\color{red}
\subsection{Training Recipes}

The generation model is trained in three stages -- token alignment, short-context pretraining, and long-context fine-tuning. We initialize the model from a pretrained Qwen3-8B checkpoint and introduce an MLP projector that maps the VQ-VAE codebook embeddings of discrete shape tokens into the Qwen hidden space. The shape-token codebook is initialized from the trained VQ-VAE and remains frozen throughout all training stages. This preserves the geometry priors learned by the VQ-VAE.
During token alignment, all Qwen parameters are frozen except the language-model head, while the MLP projector is optimized to align the newly introduced shape-token vocabulary with the pretrained Qwen representation space. This stage is trained for 1,000 iterations. The global batch size is set to 128. we use the AdamW optimizer with a peak learning rate of 1e-4 with cosine learning rate decay and linear warm-up over the first 100 iterations. During short-context pretraining, all parameters except the frozen VQ codebook are made trainable. The model is trained for 100,000 iterations on the full dataset, with each sequence truncated to at most 8,192 tokens. The optimizer configurations are the same as the first stage. Part-count balancing is not applied at this stage.
Finally, during long-context fine-tuning, the model is successively trained with maximum context lengths of 32K, 128K, and 256K tokens, for 20k, 20k and 50k iterations respectively, with maximum learning rate set to 1e-5. The corresponding global batch sizes are 128, 64 and 32. To support long context training, we set tensor parallel size to 4 and sequence parallel size to 4. Samples exceeding the corresponding maximum length are filtered out, and the part-count balancing strategy described in Supplementary Material A2 is applied.
}

%% file: subsections/a_2_ablations.tex
{\color{red}
\section{Additional Ablation Studies}
\label{sec:supp_ar_analysis}

\subsection{Object-Level Control}
\label{sec:supp_bbox_control}

We examine the effect of supervising object-level bounding-box prediction. We compare explicitly supervising the model using a object bounding box with prompting the box directly to the model for controlled generation. We compute the IoU between the user-provided object box and the actual bounding box of the generated mesh. As shown in Table~\ref{tab:supp_object_bbox}, adding object-box prediction supervision improves BBox IoU from 0.89 to 0.93, indicating stronger adherence to user-specified global constraints.

\begin{table}[t]
    \centering
    \caption{Effect of object-level bounding-box prediction supervision on global controllability.}
    \label{tab:supp_object_bbox}
    \begin{tabular}{lc}
        \toprule
        Setting & BBox IoU $\uparrow$ \\
        \midrule
        w/o object bbox prediction & 0.89 \\
        w/ object bbox prediction & \textbf{0.93} \\
        \bottomrule
    \end{tabular}
\end{table}

\subsection{Sensitivity to Part Ordering}
\label{sec:supp_part_ordering}


To investigate the sensitivity of the autoregressive generation model to the ordering of the object parts, we compare the $z$--$x$--$y$ spatial ordering used in the main paper with a connectivity-graph breadth-first-search (BFS) ordering. For $z$--$x$--$y$ ordering, part-box centers are quantized onto a $32^3$ voxel grid and sorted lexicographically, with ties broken by bounding-box volume from large to small. For BFS ordering, we construct a part-connectivity graph by voxelizing each of the components and computing the overlap, and traverse it in breadth-first order.
All other training and evaluation settings are unchanged. Table~\ref{tab:supp_part_ordering} shows comparable performance, suggesting limited sensitivity to these two reasonable serialization strategies. We retain $z$--$x$--$y$ ordering because it avoids graph construction and is faster for large-scale preprocessing.

\begin{table}[t]
    \centering
    \caption{Sensitivity to part ordering on the test set.}
    \label{tab:supp_part_ordering}
    \begin{tabular}{lcc}
        \toprule
        Ordering & CLIP $\uparrow$ & FID $\downarrow$ \\
        \midrule
        $z$--$x$--$y$ & \textbf{0.27} & 43.40 \\
        Connectivity-graph BFS & 0.25 & \textbf{41.68} \\
        \bottomrule
    \end{tabular}
\end{table}
}

%% file: subsections/a_4_more_experiments.tex
{\color{red}
\section{More Evaluation Results}
\subsection{High-Part-Count Evaluation}
\label{sec:supp_high_part}

\paragraph{Held-Out Evaluation Sets.}
To evaluate the performance of our model on dataset of varying complexity. We construct four mutually exclusive held-out test sets with ground-truth part-count ranges $[1,50)$, $[50,100)$, $[100,200)$, and $[200,300]$ of 400 objects. None of the test objects appears in the training set. For text-to-3D generation, we follow the normal-map rendering and CLIP/FID evaluation protocol used in the main paper. For bounding-box-conditioned generation, we report part Chamfer Distance (Part CD), part IoU, bounding-box IoU (BBox IoU), and success rate. Success rate is the fraction of test inputs for which inference completes and produces a valid mesh.

\paragraph{Text-to-3D Generation.}
Table~\ref{tab:supp_high_part_text} reports text-conditioned generation across the four part-count ranges. Our model maintains a CLIP score of 0.26--0.27 across all ranges. Although its FID increases from 46.3 to 73.2 as the part count grows, it remains substantially lower than all compared baselines in every range, demonstrating robust generation quality on objects with many components.

\begin{table*}[t]
    \centering
    \caption{Text-to-3D generation on held-out sets stratified by ground-truth part count.}
    \label{tab:supp_high_part_text}
    \small
    \setlength{\tabcolsep}{4pt}
    \begin{tabular}{lcccccccc}
        \toprule
        & \multicolumn{4}{c}{CLIP $\uparrow$} & \multicolumn{4}{c}{FID $\downarrow$} \\
        \cmidrule(lr){2-5}\cmidrule(lr){6-9}
        Method & $[1,50)$ & $[50,100)$ & $[100,200)$ & $[200,300]$
               & $[1,50)$ & $[50,100)$ & $[100,200)$ & $[200,300]$ \\
        \midrule
        SAR3D          & 0.24 & 0.23 & 0.22 & 0.23 & 124.0 & 118.9 & 119.9 & 116.4 \\
        Cube           & \textbf{0.26} & \textbf{0.26} & \textbf{0.27} & 0.26 & 93.5 & 99.5 & 95.0 & 100.9 \\
        TRELLIS-text   & \textbf{0.26} & \textbf{0.26} & 0.26 & 0.26 & 99.6 & 98.7 & 95.7 & 103.4 \\
        ShapeLLM-Omni  & 0.24 & 0.25 & 0.24 & 0.24 & 103.2 & 100.4 & 99.8 & 105.7 \\
        Ours           & \textbf{0.26} & \textbf{0.26} & \textbf{0.27} & \textbf{0.27}
                       & \textbf{46.3} & \textbf{55.0} & \textbf{64.9} & \textbf{73.2} \\
        \bottomrule
    \end{tabular}
\end{table*}

\paragraph{Bounding-Box-Conditioned Generation.}
For structural comparison, our model, FullPart~\cite{ding2025fullpart}, and XPart~\cite{yan2025xparthighfidelitystructure} receive the same prescribed part bounding boxes. Their auxiliary conditioning modalities differ: our model uses text, FullPart uses an image, and XPart uses a point cloud. The experiment should therefore be interpreted as a matched-layout evaluation of whether generated part geometry follows the given boxes, rather than a modality-identical comparison of unrestricted appearance quality.

Table~\ref{tab:supp_high_part_bbox} summarizes the results. Our model maintains success rates of 0.99, 0.95, 0.85, and 0.77 as the number of parts increases, while the diffusion baselines exhibit a sharp success-rate decline above 50 parts because of CUDA out-of-memory errors on the evaluation hardware. Our model also achieves consistently stronger part and box adherence. Its decreasing Part CD at larger part counts reflects that objects with many parts often contain geometrically simpler individual components, to which the adaptive representation assigns compact token budgets.

\begin{table*}[t]
    \centering
    \caption{Bounding-box-conditioned generation stratified by part count. Part CD is reported in units of $10^{-2}$, matching Table~3 of the main paper. A dash "-" indicates that no valid result is available due to OOM. Metrics other than success rate are averaged over successful outputs.}
    \label{tab:supp_high_part_bbox}
    \small
    \setlength{\tabcolsep}{6pt}
    \begin{tabular}{llcccc}
        \toprule
        Method & Part range & Part CD $\downarrow$ & Part IoU $\uparrow$ & BBox IoU $\uparrow$ & Success rate $\uparrow$ \\
        \midrule
        FullPart & $[1,50)$    & 8.3  & 0.38 & 0.66 & 0.89 \\
                 & $[50,100)$  & 10.8 & 0.26 & 0.37 & 0.17 \\
                 & $[100,200)$ & 27.2 & 0.13 & 0.33 & 0.05 \\
                 & $[200,300]$ & --   & --   & --   & 0.00 \\
        \midrule
        XPart    & $[1,50)$    & 7.7  & 0.45 & 0.42 & 0.82 \\
                 & $[50,100)$  & --   & --   & --   & 0.00 \\
                 & $[100,200)$ & --   & --   & --   & 0.00 \\
                 & $[200,300]$ & --   & --   & --   & 0.00 \\
        \midrule
        Ours     & $[1,50)$    & \textbf{1.9} & \textbf{0.68} & \textbf{0.93} & \textbf{0.99} \\
                 & $[50,100)$  & \textbf{1.0} & \textbf{0.72} & \textbf{0.90} & \textbf{0.95} \\
                 & $[100,200)$ & \textbf{0.6} & \textbf{0.76} & \textbf{0.92} & \textbf{0.85} \\
                 & $[200,300]$ & \textbf{0.2} & \textbf{0.81} & \textbf{0.87} & \textbf{0.77} \\
        \bottomrule
    \end{tabular}
\end{table*}

All methods are evaluated on a single NVIDIA A800 80~GB GPU. Note that our model can also fail for exceptionally long sequences under this memory budget. This limitation is governed primarily by sequence length rather than part count. Some objects with fewer than 300 parts require more tokens than some 300-part objects, depending on the part complexity. Using additional GPU memory or distributed inference can extend the feasible context length.

\subsection{Inference Efficiency and Memory}
\label{sec:supp_efficiency}

We profile single-object inference on one NVIDIA A800 80~GB GPU using bfloat16 precision. Autoregressive generation is implemented with Hugging Face Transformers, and shape extraction decodes each part sequentially before applying marching cubes at resolution $1024^3$.
Table~\ref{tab:supp_inference_time} separates autoregressive generation from VQ-VAE decoding and marching cubes.

\begin{table}[t]
    \centering
    \small
    \resizebox{\linewidth}{!}{
    \begin{tabular}{lrrrr}
        \toprule
        Part range & $[1,50)$ & $[50,100)$ & $[100,200)$ & $[200,300]$ \\
        \midrule
        AR generation & 335 & 2066 & 3746 & 6735 \\
        Decoding + marching cubes & 51 & 413 & 502 & 1076 \\
        \bottomrule
    \end{tabular}
    }
    \caption{Inference-time breakdown in seconds across part-count ranges.}
    \label{tab:supp_inference_time}
    \vspace{-6mm}
\end{table}

Naive Hugging Face generation achieves an average throughput of 17.0 tokens/s, confirming that autoregressive decoding is the primary bottleneck. Because the generator is based on Qwen3, standard LLM serving systems can be used directly: replacing the naive implementation with vLLM increases single-GPU throughput to 42.5 tokens/s. Model quantization, speculative or multi-token decoding, and distributed inference are compatible directions for further acceleration; claims about a specific additional throughput should be reported only with a measured configuration.

Mesh extraction introduces additional overhead because all parts are decoded sequentially and marching cubes is evaluated at $1024^3$ resolution for every part, even when the part geometry is simple. This cost can be reduced by lowering the extraction resolution, batching part decoding, or overlapping shape decoding with autoregressive generation.

For one generated 300-part object with a context length of 188k tokens, peak GPU memory is 71.24~GB, which fits on a single 80~GB GPU. This example should not be interpreted as a fixed memory cost for all 300-part objects: peak memory is determined mainly by token length, and part count is not strictly monotonic with sequence length.

\subsection{Rate--Distortion Selection Cost}
\label{sec:supp_rd_cost}

Rate--distortion selection is an offline preprocessing step used only to construct adaptive-length training sequences for the LLM. For each part, we sweep nine candidate prefix lengths,
$L\in\{2^n\}_{n=4}^{12}$, and evaluate the candidates with GPU parallelism. The complete sweep takes 1.8 seconds per part on average. It introduces no additional generation-time overhead because the autoregressive model directly predicts the number of shape tokens used for each part.

\subsection{Human Perceptual Evaluation}
\label{sec:supp_user_study}

We conduct a human perceptual study to assess (i) alignment between the text prompt and generated geometry and (ii) overall geometry quality. We sample 20 prompts and generate one result per method. Each result is rendered from 6 fixed viewpoints using identical camera and lighting settings. The method order is randomly shuffled and the identities are hidden. The preference score is computed as the fraction of votes received by each method.

As shown in Table~\ref{tab:supp_user_study}, our model receives 64\% of the votes for text alignment and 69\% for geometry quality, substantially exceeding the compared methods.

\begin{table}[th]
    \centering
    \caption{User studies for text alignment and geometry quality.}
    \label{tab:supp_user_study}
    \small
    \begin{tabular}{lcc}
        \toprule
        Method & Text alignment $\uparrow$ & Geometry quality $\uparrow$ \\
        \midrule
        SAR3D         & 0.00 & 0.00 \\
        Cube          & 0.18 & 0.16 \\
        TRELLIS-text  & 0.16 & 0.13 \\
        ShapeLLM-Omni & 0.02 & 0.02 \\
        Ours          & \textbf{0.64} & \textbf{0.69} \\
        \bottomrule
    \end{tabular}
\end{table}

}

%% file: subsections/a_3_model_capability.tex
{\color{red}\section{Additional Visualization Results}}
{\color{red}\subsection{Additional Emergent Generation Behaviors}}
\label{sec:appendix_emergent_generation}

\begin{figure}[tb]
  \centering
  \includegraphics[width=\linewidth]{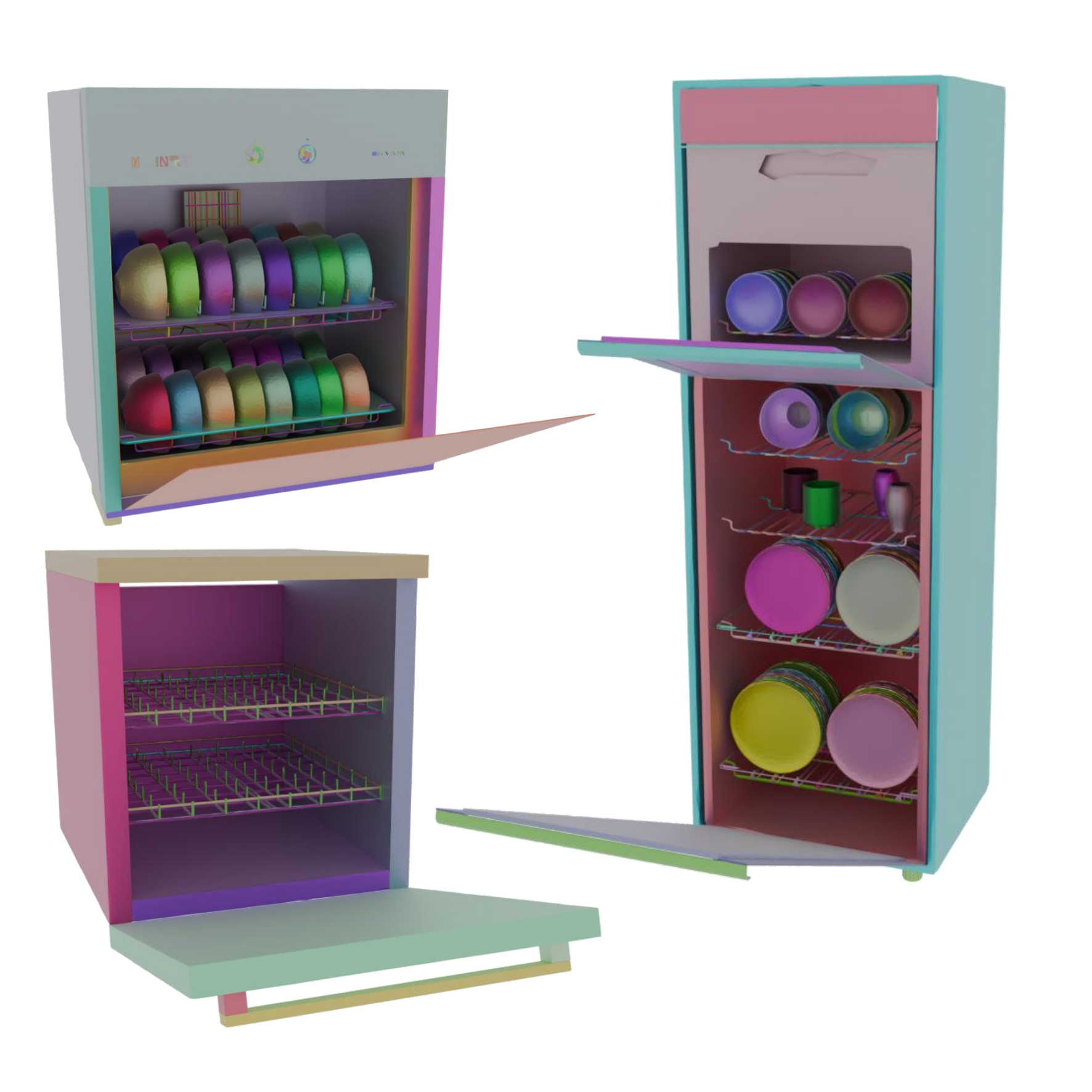}
  \caption{Three dishwashers with high quality interior details generated by our model.}
  \label{fig:appendix_emergent_cases_1}
\end{figure}

Although our model is not explicitly trained for interior-structure modeling, we observe that the model occasionally produces objects with plausible internal structures. Representative examples are shown in Figure~\ref{fig:appendix_emergent_cases_1}.

\begin{figure*}[htp]
  \centering
  \includegraphics[width=0.87\linewidth]{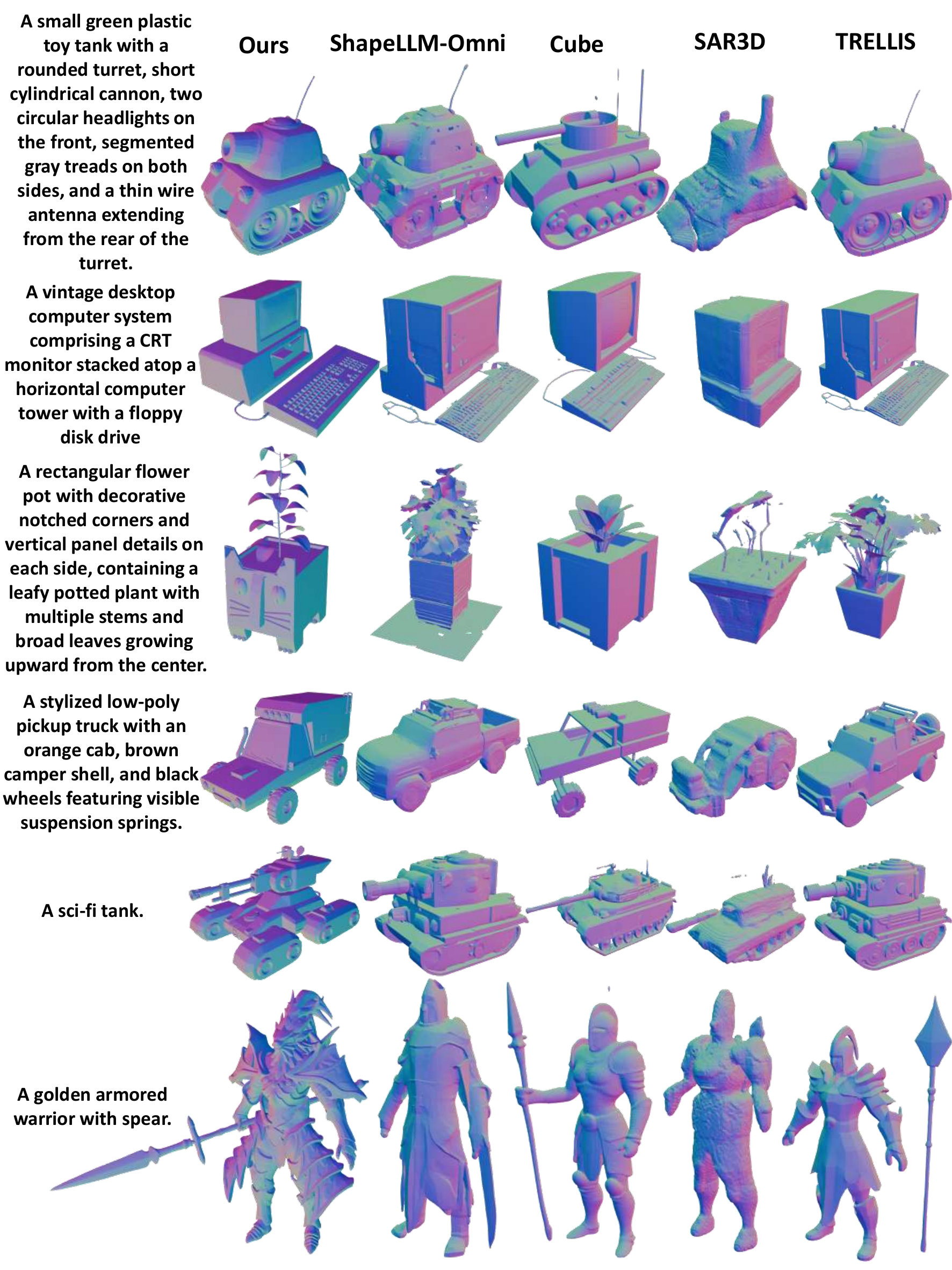}
  \caption{More visual comparisons between our model and baselines on text conditioned 3D generation.}
  \label{fig:appendix_emergent_cases_2}
\end{figure*}

{\color{red}
\subsection{Additional Text to 3D Generation Results}
We present additional visualization results on text to 3D generation in Figure ~\ref{fig:appendix_emergent_cases_2}.
}

